\documentclass[journal,comsoc]{IEEEtran}
\usepackage[T1]{fontenc}% optional T1 font encoding
\usepackage{graphicx}
\usepackage{times}
\usepackage{helvet}
\usepackage{courier}
\usepackage{amsmath}
\usepackage{algorithm}
\usepackage{algorithmic}
\usepackage{csquotes} 
\usepackage{color}
\usepackage{paralist}
\usepackage{amssymb}
\usepackage{indentfirst}
\usepackage{subfigure}
\usepackage{float}
\usepackage{multirow}
\usepackage{cite}
\usepackage{hyperref}
\usepackage{mathrsfs}
\usepackage{amsthm}

\newtheorem{defn}{Definition}%[section]

\usepackage{cite}
\ifCLASSINFOpdf
\else
\fi

\usepackage{amsmath}
\usepackage[cmintegrals]{newtxmath}

\begin{document}

\title{Deep Co-Space: Sample Mining Across Feature Transformation for Semi-Supervised Learning}

\author{Ziliang Chen, Keze Wang, Xiao Wang, Pai Peng, Ebroul Izquierdo, and Liang Lin
\IEEEcompsocitemizethanks{This work was supported in part by the
State Key Development Program under Grant 2016YFB1001004, in part by
the National Natural Science Foundation of China under Grant 61671182, in part by Hong Kong Scholars Program and Hong Kong Polytechnic University Mainland University Joint Supervision Scheme, and sponsored by CCF-Tencent Open Research Fund (NO. AGR20160118). Corresponding author is Liang Lin.
\IEEEcompsocthanksitem Z. Chen, K. Wang and L. Lin are with the School of Data and Computer Science, Sun Yat-sen University, Guangzhou, China and also with Engineering Research Center for Advanced Computing Engineering Software of Ministry of Education, China. Email: zlchilam@163.com; kezewang@gmail.com; linliang@ieee.org. Keze Wang is also with Dept. of Computing, The Hong Kong Polytechnic University, Hong Kong.
\IEEEcompsocthanksitem X. Wang is with the School of Computer Science, Anhui University, Hefei, P. R. China. Email: wangxiaocvpr@foxmail.com.
\IEEEcompsocthanksitem P. Peng is with Youtu Lab of Tencent, Shanghai, P. R. China. Email: popeyepeng@tencent.com.
\IEEEcompsocthanksitem E. Izquierdo is with the School of Electronic Engineering and Computer Science, Queen Mary University of London, London, U.K. (e-mail: ebroul.izquierdo@qmul.ac.uk).}}

\markboth{IEEE Transactions Circuits and Systems for Video Technology}%
{Chen \MakeLowercase{\textit{et al.}}: Deep Co-Space: Sample Mining Across Feature Transformation for Semi-Supervised Learning}

\maketitle

\begin{abstract} 
	Aiming at improving performance of visual classification in a cost-effective manner, this paper proposes an incremental semi-supervised learning paradigm called Deep Co-Space (DCS). Unlike many conventional semi-supervised learning methods usually performing within a fixed feature space, our DCS gradually propagates information from labeled samples to unlabeled ones along with deep feature learning. We regard deep feature learning as a series of steps pursuing feature transformation, i.e., projecting the samples from a previous space into a new one, which tends to select the reliable unlabeled samples with respect to this setting. Specifically, for each unlabeled image instance, we measure its reliability by calculating the category variations of feature transformation from two different neighborhood variation perspectives, and merged them into an unified sample mining criterion deriving from Hellinger distance. Then, those samples keeping stable correlation to their neighboring samples (i.e., having small category variation in distribution) across the successive feature space transformation, are automatically received labels and incorporated into the model for incrementally training in terms of classification. Our extensive experiments on standard image classification benchmarks (e.g., Caltech-256~\cite{Griffin2007Caltech} and SUN-397~\cite{Xiao2014SUN}) demonstrate that the proposed framework is capable of effectively mining from large-scale unlabeled images, which boosts image classification performance and achieves promising results compared to other semi-supervised learning methods.
\end{abstract}

\IEEEpeerreviewmaketitle

% ------------------------------------------------------------------------------------- Introduction 
\section{Introduction}
\label{intr}
\IEEEPARstart{R}{ecently}, tremendous advancements have been made in the field of vision by convolutional neural networks (CNNs), including classification~\cite{Krizhevsky2012ImageNet}, object detection~\cite{girshick2014rich}, scene and human parsing~\cite{lin2016deep, liang2015deep} and image caption generation~\cite{vinyals2015show}. The successes on these vision applications have exhibited impressive performances with ample well-annotated images for training. Though label information plays such a crucial role in those applications, the establishment of large scale dataset is too expensive to be affordable under a practical scenario. Besides, annotation by human labor also brings about labels contamination caused by the limitation of knowledge background from the ordinary workers. 

As the growing demand of improving the usage of existing label information to reduce the annotation cost~\cite{wang2016cost}~\cite{wanglearning}, semi-supervised learning (SSL) obtains increasing attention. By ingeniously bridging the connection among unlabeled data and labeled information, SSL performs well with a limited number of labeled samples. This cost-effective property makes SSL always in the forefront of computer vision and machine learning research. Currently, the progress of deep learning focuses on two branches for SSL algorithms, i.e., feature-fixed and feature-learnable SSL. The former usually refers to a variety of conventional SSLs (e.g., Graph-based SSL\cite{Zhu2008Semi, goldberg2009multi, ding2011learning}), which consider samples in a handcrafted feature space during the whole training process. Differently, the latter additionally focuses on learning representation according to SSL configuration. Through learning both feature representation and training model parameter simultaneously, this branch usually pays close attention to the exploration about nonlinear functions approximation via semi-supervised metric learning~\cite{yu2012semisupervised} and newly rising deep learning~\cite{deng2012mnist, krizhevsky2014cifar}.

%Under the smootheness assumptions in feature space, graph-based learning machine is trained along with graph-based local data relationship. Aided by well-defined and low-dimensional features, GSSL can achieve competitive performance even if only a few labeled samples are provided in training. 
% 

\begin{figure*}[t]
	\center
	\includegraphics[width=7in]{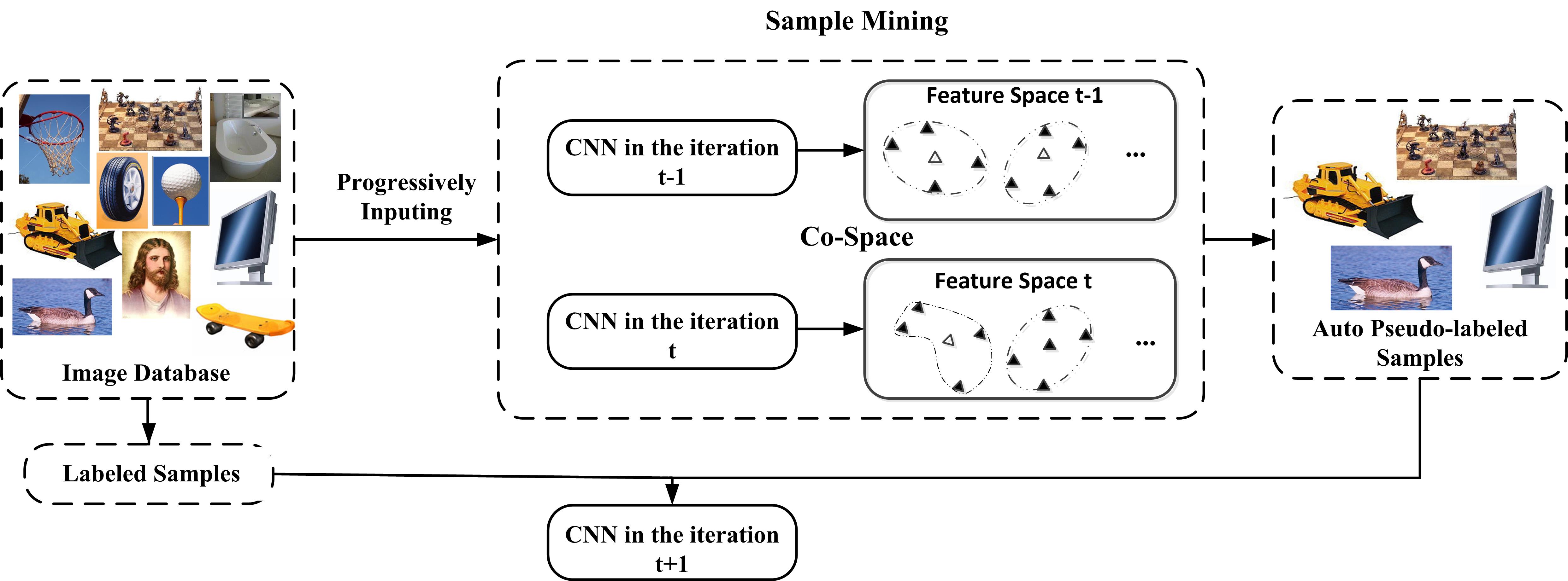}
	\caption {The pipeline of the proposed Deep Co-Space framework. At the beginning, we have limited labeled data and infinite unlabeled data for training. The labeled data would be used to fine-tune a pre-trained CNN-based deep model and results in a new one. After that, all labeled and unlabeled data will be extracted by the old and new models to construct two successive feature space (Co-Space) respectively. We measure the distribution variation of labeled neighbors for each unlabeled sample in Co-Space, and assign those samples having stable structures with pseudo-labels. Then these selected samples are employed to update the model for the next iteration.}
	\label{pipeline}
\end{figure*}

In spite of achieving remarkable successes in visual recognition, these two branches still face several limitations. In specific, conventional feature-fixed SSLs heavily rely on the feature engineering, which tends to  
strengthen some information illustrated in statistics and discard other information in visual aspect as a return. This leads to its failure under task-orientated scenario, which requires accurate feature representation to be adaptive to different visual understanding tasks. In respect to feature-learnable SSLs, though seeking overall distribution in feature space~\cite{zhao2015stacked} under an end-to-end network training regime, it cannot be progressively optimized in such an incremental way due to the ignorance of modeling the local relationship among samples\cite{kipf2016semi}. As discussed in~\cite{weston2012deep, li2015towards}, these aforementioned limitations are still arousing wide concern in research.

%Liang~\cite{} formulated an incremental semi-supervised learning framework to train a network-based object detector via transferring knowleadge from video.

%Attempting at overcoming the non-compatibility problem, Weston invent deep semi-supervised embedding (DSSE) ~\cite{weston2012deep} in the persperctive of training neural network with graph-based regularity. Incorporating the graph relationship as a balancing loss into parameter updating, DSSE receives positive results with different configurations of network structure in many semi-supervised classification tasks. The approach can be treated as a GSSL method in deep learning version. Nevertheless, the relational graph describing locality among data, must be pre-defined before training. This prerequisition is common in social network analysis\cite{keylist}. However, in visual classification problem, delivering a total relation information among all samples is actually a strong supervision and not a usual case.

%The drawback motivates us to develop a novel SSL model that explores local data relationship by the model itself.

Attempting to overcome these limitations from another point of view, we introduce an innovative sample mining strategy, which incrementally explores the related local structure for each unlabeled sample within two different feature spaces. More specifically, assuming that each couple of deep learning models being fine-tuned before/after can be viewed as two successive yet different feature spaces, we define the one-one nonlinear correspondence for each sample from the previous feature space to a new one as ``feature transformation''. As feature space transforming, according to low density separability assumption\cite{Zhu2005Semi}, samples in the same category tend to cluster together, keeping locally compact and semantically coherent, then those in different classes are inclined to diverge to corresponding categories, which leads to its labeled neighbors apparently to change. More precisely, with regard to each unlabeled sample in the transformed feature space, its labeled neighbors tend to form a locally stable distribution for a certain category. This inspires us that some unlabeled samples, remaining stable labeled neighbor distribution in the two successive feature spaces, can be employed by assigning them pseudo-labels to augment labeled dataset and improve the performance. 

As illustrated in Fig.~\ref{pipeline}, an innovative incremental sample mining framework based on the intuition above, is proposed for deep semi-supervised learning. Since the progressive sample mining seems like a sequence of steps pursuing feature space transformation via gradually polishing a deep model, we name our framework as \textbf{Deep Co-Space (DCS)}, namely, there are two CNN models with the same architecture at each step in our framework. Note that, the second CNN has been fine-tuned based on the first one via the updated labeled data pool, then the training phase is performed as follows. Firstly, we extract features for all labeled and unlabeled samples based on these two CNN models, thus, we have obtained two successive feature spaces (Co-Space); Secondly, we launch our sample mining strategy to select those unlabeled samples keeping local stable neighbor distributions in the Co-Space and automatically annotate them with pseudo-labels. More specifically, for each unlabeled sample, we measure its reliability by calculating the category variations of its $K$ nearest neighbors in Co-Space. In order to resist semantic drift~\cite{curran2007minimising}, the variation is required under consideration of two points of views, i.e., neighborhood intrinsic variation and neighborhood category variation. Neighborhood intrinsic variation represents the intrinsic structure non-consistency of unlabeled samples via feature transformation in the Co-Space, while 
neighborhood category variation denotes the transforming variation of covariances among local labeled samples related to different categories. Then, we merge them into an unified sample mining criterion, which is based on Hellinger distance.  
Finally, given samples selected by this criterion, we augment labeled data pool in the image database and further fine-tune the CNNs. In this way, the updated CNN leads to a new Co-Space for sample mining at the next iteration.

The main contributions of this paper are in three-fold: i) To the best of our knowledge, DCS is the first incremental semi-supervised learning framework attempting to progressively propagate information from labeled samples to unlabeled ones along with a sequence of steps, which aims at leveraging feature transformation in a two successive feature space; ii) We present sufficient discussions and clarifications about how to incorporate the neighborhood intrinsic and category variation into an unified sample mining criterion deriving from Hellinger distance; iii) Extensive experiments on two public visual classification benchmarks, i.e., Caltech-256\cite{Griffin2007Caltech} and SUN-397 \cite{Xiao2014SUN}, demonstrate the effectiveness of our DCS in SSL not only on the vanilla Alexnet\cite{Krizhevsky2012ImageNet} and VGG\cite{simonyan2014very}, but also on the recent network architectures\cite{lee2013pseudo}\cite{zhao2015stacked} with the deep semi-supervised learning losses.

The rest of the paper is organized as follows. Sect.~\ref{related}
presents a review of approaches related to DCS. Sect.~\ref{dcs} overviews
the complete model about DCS, including definition, pipeline, and some theoretical discussion. The
experimental results, comparisons and component analysis are
presented in Sect.~\ref{exper}. Finally, Sect.~\ref{conclusion} concludes the paper. %------------------------------------------------------------------------------------- Related Work 

\section{Related Work} 
\label{related}
In this section, we give a brief review of some feature-fixed SSL approaches related to our framework, and the SSL methods related to neural network are exhibited in Sect.~\ref{ssl}. Since our DCS shares some properties of multi-view learning, then the comparability and difference between them are discussed in Sect.~\ref{mvl}.
\subsection{Semi-supervised Learning}
\label{ssl}
\subsubsection{Feature-fixed SSL}
The most aged SSL method starts from self-training\cite{Yarowsky1995Unsupervised}, which trains a classifier with small amount of labeled data to annotate unlabeled data, then retrain the classifier with labeled and unlabeled data iteratively. Self-training is straightforward both in intuition and formulation, but always beset by semantic drift. It has been extended into many variants\cite{Nigam2015Analyzing}\cite{chen2013neil} to prevent this problem, and most of them rely on knowledge from fixed feature space. 

Probabilistic graphical model plays an important role in the development of SSL. For instance, Ji et al.~\cite{zhang2005semi} merges the supervised and unsupervised hidden Markov models into an associated estimation problem as a set of fixed point equations; Mao et al.~\cite{mao2012sshlda} explores new latent topic in LDA (Latent Dirichlet Allocation) with labeled hierarchical information. All of them utilize all data to model the joint probability distribution in generative process with discriminative information. They are well-defined in theory, but suffer from high variance in generative process when the assumption of prior distribution is inappropriate. Besides, compared with deep learning, pure graphical models rely on features with high-level semantics in statistics, which makes those methods more preferable in addressing problems about natural language processing.

Graph based semi-supervised learning draws attention of many researchers both in transductive and inductive learning settings, such as label propagation~\cite{Zhu2005Semi}, manifold regularization~\cite{belkin2006manifold}, Planetoid~\cite{yang2016revisiting} e.t.c. The problem is usually formulated as
\begin{displaymath}
\label{h1}
\underset{f}{min} \ \ \\ \lambda f(X)^T \Delta f(X) + \mathcal{L}(f(X), \ Y)
\end{displaymath}
where $\Delta =  \begin{bmatrix} L_{uu} & L_{ul} \\ L^{T}_{ul} &L_{ll}  \end{bmatrix}$ is a matrix about unlabeled and labeled dataset, and is related to the finite weighted graph $\mathcal{G} = (V,E,W)$. Specifically, $\mathcal{G}$ consists of a set of vertexes $V$ based on all data, and can be provided from external knowledge or pre-definition. The edge set $E$ and its specified weights $W$ are formulated with non-negative symmetric function. Note that $\Delta$ is also determined before optimizing. When $\mathcal{G}$ is required for calculation, we will interpret the $W(i,j)$ as a local similarity measure between the vertexes $x_i$ and $x_j$. Then based on $K$ nearest neighbor graph ($K$nn), the element of weighted matrix $\Delta$ is denoted as:

\begin{equation}
\label{graph}
\Delta(i,j) = \frac{W(i,j)}{\sum_{x_k\in Knn(x_i)}W(i,k)}
\end{equation}

\begin{displaymath}
s.t. \ \ W(i,j) = \left\{
\begin{aligned}
& \ \frac{h(\frac{\rho(x_i,x_j)^2}{\mu\sigma^2})}{\underset{x_k\in Knn(x_i)}{\sum} h(\frac{\rho(x_i,x_k)^2}{\mu\sigma^2})}  \ & x_j\in Knn(x_i)\\
& \ 0 \ & otherwise\\
\end{aligned}
\right.
\end{displaymath}
where $h$ is a function with exponential decay at infinity, which is often $exp(-x)$. $\rho$ is a distance measurement between two given samples. $\mu$ and $\delta$ are both hyper-parameters. Moreover, $\delta$ can be calculated by mean distance to $Knn$ of $x_i$\cite{wang2013dynamic}. In the case of transductive learning, $f$ is always denoted as: $\begin{bmatrix}
\textbf{f}_u \\ \textbf{f}_l
\end{bmatrix}$, in which $\textbf{f}_l$ and $\textbf{f}_u$ are label probabilities for labeled and unlabeled data respectively.

%Then the distribution of point $x$ in transductive GSSL is obtained as:
%\begin{displaymath}
%	\label{h2}
%	\textbf{f}_u = -L^{-1}_{uu}L_{ul}\textbf{f}_l
%\end{displaymath}
%where $\textbf{f}_l$ is always replaced by the probablistic form of $Y$.

As the Sect.~\ref{intr} exhibits, DCS aims at searching unlabeled samples that have kept stable correlations with its neighbors during feature space transformation by measuring intrinsic variation and category variation. Transductive GSSL is an ideal bridge to estimate the intrinsic structure among unlabeled samples. In the implementation of our DCS, we employ label propagation for transductive label inference (Please see more details in Sect.~\ref{dcs}.

\subsubsection{Feature-learnable SSL (DSSL)} DSSL for visual classification is usually categorized into two classes: reconstruction model and generation model. The former focuses on training deep model with reconstruction architecture in SSL manner~\cite{zhao2015stacked}. It presents as a mirror architecture with encoding and decoding pathways like auto-encoder, and makes discrimination and unsupervised reconstruction for all data during the training phase. On the contrary, generation model achieves semi-supervised learning through creating data to classify. The generation model based methods start from deep generative network~\cite{kingma2014semi}, and have received a great success with the development of generative adversarial network (GAN)~\cite{radford2015unsupervised, zhao2016energy} and variational auto-encoder~\cite{kingma2013auto} . In fact, the idea of generation model is close to semi-supervised graphical model as we have mentioned above. In other words, both of them make inference and generation with discrimination. However, unlike pure graphical model, generation model shows more promising in generating data in continuous space (image and video).

Some researchers focus on combining neural networks and conventional methods. Liang et al.~\cite{liang2015towards} formulated an incremental semi-supervised learning framework to train a network-based object detector via transferring knowledge from video.
The incremental active learning technique by Lin~\cite{lin2017active}, achieving an cost-effective labor in manual labeling, has recently received great attentions in the deep CNNs area for visual recognition. Weston et al.~\cite{weston2012deep} invented deep semi-supervised embedding (DSSE) in the perspective of training neural network with graph-based regularity. Incorporating the graph relationship as a balancing loss into parameter updating, DSSE receives positive results with different configurations of network structure in many semi-supervised classification tasks. The approach can be treated as a GSSL variant in deep learning. Nevertheless, the relational graph describing locality among data, must be pre-defined before training. This premise is common in social network analysis~\cite{kashima2009link}. However, in visual classification problem, delivering total relation information among all samples is actually a strong supervision and not an usual case.

%	DCS is known as a data-driven semi-supervised sample selection strategy base on deep learning and also merge with traditional method. It runs on a CNN being fine-tuning, and improve the network performance in semi-supervised learning scenario. 

\subsection{Mult-view Learning (MVL)}
\label{mvl}
Multi-view learning is a family of learning algorithms deriving from Co-Training~\cite{Nigam2015Analyzing}, and focuses on exploiting data with multi-representation. For example, a cartoon character can be represented by different views of the character like color histogram, skeleton and contour~\cite{yu2012semisupervised}, and the views facilitate select reliable samples and label them in supplementary way. The recent MVLs have extended to many kinds of application, e.g., clustering~\cite{kumar2011co}, reconstruction~\cite{xu2015multi} and representation learning~\cite{su2015multi}. It is interesting that the design about Co-Space is similar to two views learning, yet where both views come from the features extracted by the networks before and after fine-tuning respectively. Both of them make the decision about labeling according to both views together. Nevertheless, MVL usually results in two classifiers, which are implemented in the SSL case. In contrast, DCS proposes to perform sample mining via feature transforming in the chain of Co-Spaces. Those Co-Spaces are built upon a single neural network with different parameters, which are obtained via fine-tuning a network with an image database incrementally. Besides, in MVL setting, each view has been assumed to be independent to other views~\cite{xu2013survey}, but both views in Co-Space apparently correlate with each other in some way instead.

%\subsection{Self-paced learning (SPL)}
%\label{spl}
%Inspired by the cognitive science result, self-paced learning~\cite{Kumar2010self, jiang2015self} is a recently proposed model training regime, which incorporates training samples from "easy" to "hard" and solve machine learning problem by alternating convex optimization. In each iteration, the weight for each sample is assigned as cost-sensitive coefficient and ranges from $0$ to $1$. And if the weight only chosen from $\{0,1\}$ , the SPL framework can be degenerated to a sample mining method to solve a subset selection problem.  "Easy" sample which has smaller loss in the model will be assigned larger weight, and the "hard" vice versa. Then as the training paradigm aging, the weight about hard sample gradually becomes more influential.  
%
%DCS processes training samples iteration by iteration as SPL, and considers more on the aspect from "reliable" to "unreliable" instead of from "easy" to "hard". Through this incremental manner, DCS prevents some semantic drift, which is common in pseudo labeling method. 

% --------------------------------------------------------------------------------------------------------------------------------------------------------------------------------------------------------------
% ------------------------------------------------------------------------------------- The Proposed Approach -
\section{Deep Co-Space}
\label{dcs}
In this section, we discuss the formulation of our proposed framework. In Sect.~\ref{31}, we introduce the pipeline of DCS, then the concept about Co-Space and feature transformation are defined.  We leverage the feature transformation to formulate the sample mining strategy in Sect.~\ref{32}, which is most important part in DCS. Finally, further analysis about the strategy is discussed in Sect.~\ref{33}.

\subsection{DCS architecture and Co-Space}
\label{31}
In the context of visual classification, suppose that we have $n$ samples taken from $m$ classes for training. They are raw image data and we denotes the image database as $D = \{$\textbf{x}$_i\}_{i = 1}^{n}$. Then one-hot vector $\mathbf{y}_i = \{y_k\}_{k = 1}^{m}$ represents the label for \textbf{x}$_i$ and the $\mathbf{Y}$ is category set. In the setting of semi-supervised learning, only parts of images in $D$ are labeled. For the simplicity in further discussion, we denote $D^L$ as labeled images and $D^U$ as unlabeled images respectively. 

%Further more, in the case of incremental learning, $D^L$ will be gradually enriched with reliable pesudo-labeled images, which means the propotion of labeled images in $D$ also dynamically change. Correspondingly, we denotes the initiation of $D$ as $D_0$.

A CNN-based model $f_{\theta}$ is introduced to attain visual classifier and deep feature learning jointly, which $f$ is the network architecture and $\theta$ means its parameter. The CNN model has been pre-trained by some large scale visual recognition database, which contributes some of visual semantics to the initial $f$. As the description about DCS in Sect.~\ref{intr}, feature for each image is extracted iteratively and utilized to calculate the category consistency in feature transformation. Then the feature for image \textbf{x}, which extracted from the network $f$ in iteration $t$, is denoted as $f_{\theta_t}($\textbf{x}$)$. (The output of \textbf{x} in $f_{\theta}$ is a result of classification. Since we don't use the classification result to explore sample in DCS, the $f_{\theta}($\textbf{x}$)$ is treated as the output feature map\textbf{/}vector for \textbf{x} which extracted from $f_{\theta}$ in our formulation.)

We use $\theta_0$ to present the pre-trained $f$ parameter, and after the fine-tuning with $D$, $f_{\theta_0}$ leads to an updated model $f_{\theta_1}$ in the first iteration. In analogy, $f_{\theta_t}$ is the updated model in the $t$-th iteration, which have been fine-tuned from the model $f_{\theta_{t-1}}$ with renewed dataset augmented by sample mining result in previous iteration. Here we 
obtain the definition of Co-Space as follows:

\begin{defn}(\textbf{Co-Space})
	Suppose $f_{\theta}(D)$ is a feature set for dataset $D$, which extracted from model $f_{\theta}$. The couple of feature sets <$f_{\theta_{t-1}}(D)$, $f_{\theta_{t}}(D)$>, is defined as the Co-Space of dataset $D$ in the iteration $t$.
\end{defn}

\begin{figure}[t]
	\center
	\includegraphics[width=2.4in]{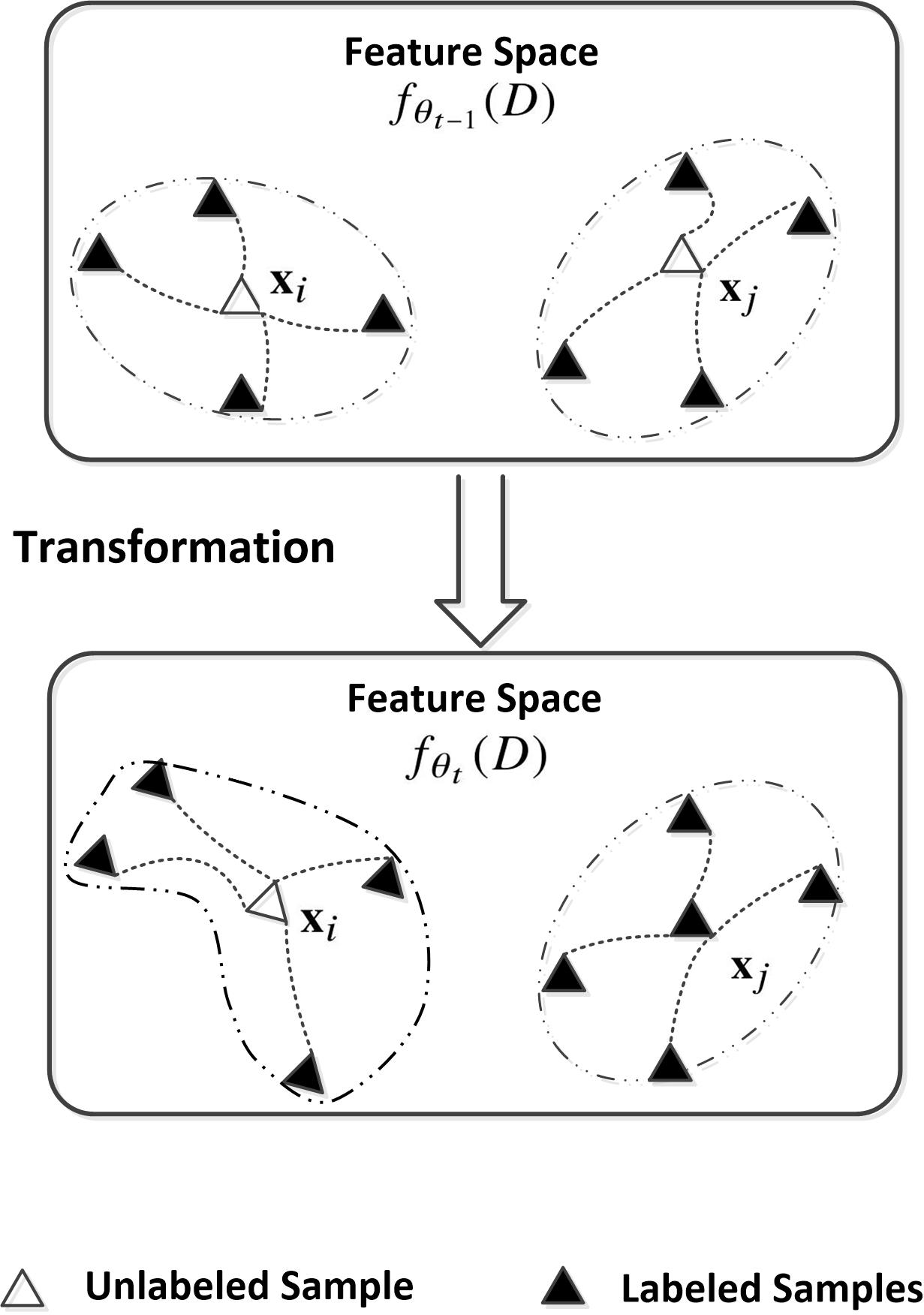}
	\caption {We consider unlabeled instance $\mathbf{x}_i$ and $\mathbf{x}_j$ according to their neihborhoods. Area in dotted lines denotes the intrinsic structure around them. As we can see, instance $\mathbf{x}_i$ changes a lot in intrinsic structure, and its labeled neighbors also vary in transformation; and vice versa for instance $\mathbf{x}_j$. Eventually, instance $\mathbf{x}_j$ is selected and labeled.}
	\label{tu1}
\end{figure}

 Notice that in the definition above, Co-Space is generally interpreted as the construction in the $t$-th iteration. When $t$ equals $1$, a Co-Space is obtained as the description above; as DCS works, a set of unlabeled samples will be selected as pseudo-labeled candidates and used to retrain the $f$ from $\theta_1$ to $\theta_2$, which leads to the next Co-Space <$f_{\theta_{1}}(D)$, $f_{\theta_{2}}(D)$>, and then  following this process of deduction. As we notice, dataset $D$ is non-specific, which means $D$ also representing any subset in the whole possible data space. 
 
 \begin{algorithm}[htb]
	\caption{ Label Propagation with $K$nn \cite{Zhu2005Semi}  }
	\label{Algorithm1}
	\begin{algorithmic}[1]
		
		\REQUIRE ~~\\
		
		Labeled dataset $D^L$ and unlabeled dataset $D^U$; The label set $Y^L$ corresponding to labeled dataset $D^L$; $\delta$; $\mu$; max iteration $T$.\\
		
		\ENSURE ~~\\
		Soft labels $Y^U$ for unlabeled dataset.
		
		\STATE Utilize Eq.~(\ref{graph}) to initiate transition matrix $P$ with $D_L$, $D_U$, $\delta$ and $\mu$.
		\STATE Initiate soft label set $Y_0 = [Y_0^L;Y_0^U]$, which $Y^U = \ $\textbf{0}.
		
		\FOR { t = 1 to $T$}
		
		\STATE $[Y^L_t;Y^U_t] = P*[Y_{t-1};Y_{t-1}]$.
		
		\STATE $Y_t^L = Y^L$.
		
		\ENDFOR		
		
		\STATE $Y^U = Y^U_T$.
		
	\end{algorithmic}
\end{algorithm}
 
 Using Co-Space, the feature transformation is defined as below:
 
\begin{defn}(\textbf{Feature Transformation})
	Provided dataset $D$, <$f_{\theta_{t-1}}(D)$, $f_{\theta_{t}}(D)$> is a corresponding Co-Space in the iteration $t$. For all $x$ belong to $D$, the projection $F^D_t: f_{\theta_{t-1}}(x) \rightarrow f_{\theta_{t}}(x)$ denotes the feature space transformation in the iteration $t$; and given specific sample \textbf{x}, the feature transformation denotes as $F_t(\mathbf{x}) = <f_{\theta_{t-1}}(\mathbf{x}),  f_{\theta_{t}}(\mathbf{x})>$.
\end{defn}

Obviously, $F_t($\textbf{x}$)$ is an one-to-one matching feature relation in Co-Space, thus for each \textbf{x}, there is only one $F_t($\textbf{x}$)$ acting as response. As for each unlabeled sample \textbf{x}, DCS launches sample selection according to $F_t($\textbf{x}$)$, which promising that given an unlabeled sample \textbf{x}, there is single one decision in the sample mining.

Both of the definitions compose the basis of our sample mining strategy. As an overview in brief, Fig. \ref{tu1} demonstrates how to select reliable samples via feature transforming in Co-Space. In specific, sample $\mathbf{x}_i$ and $\mathbf{x}_j$ are both unlabeled samples we considering to select. In the definition, they have two feature expressions in Co-Space, corresponding to their feature transforming respectively. After the transformation, as for $\mathbf{x}_i$, the correlation between $\mathbf{x}_i$ and its neighbors largely change ( intrinsic structure varies as the change of unlabeled neighbors; labeled local sample covariance varies as the change of labeled neighbors ). In contrast, the category of $\mathbf{x}_j$ and its local samples keeps relatively stable. According to this leaked information, $\mathbf{x}_j$ is more preferable as a reliable candidate.

Specifically in each iteration, we use initial training set or training set augmented by pseudo-labeled sample, to update the model and obtain a new Co-Space. The Co-Space brings about feature transforming in dataset, which leads to the sample selection decision for each unlabeled samples. Then those selected samples are plugged into labeled data pool to replay the progressively semi-supervised learning process.

 \begin{algorithm}[htb]
	\caption{ Neighborhood Intrinsic Variation  }
	\label{Algorithm2}
	\begin{algorithmic}[1]
		
		\REQUIRE ~~\\
		
		Labeled Co-Space <$f_{\theta_b}(D^L),f_{\theta_a}(D^L)$> and unlabeled Co-Space <$f_{\theta_b}(D^U),f_{\theta_a}(D^U)$>, where $\theta_{b}$ and $\theta_{a}$ corresponding to mode parameter before/after updating; The label set $Y^L$ corresponding to Co-Space <$f_{\theta_b}(D^L),f_{\theta_a}(D^L)$>; $\delta$; $\mu$; max iteration $T$.
		
		\ENSURE ~~\\
		Soft labels $Y_{\theta_b}^U$ and $Y_{\theta_a}^U$ for unlabeled dataset; low dimensional feature sets $f_{\theta_b}(D)$ and $f_{\theta_a}(D)$ for $D = D^U\cup D^L$. \\
		\STATE Obtain $f_{\theta_b}(D)$ = $f_{\theta_b}(D^L)\cup f_{\theta_b}(D^U)$.
		\STATE Obtain $f_{\theta_a}(D)$ = $f_{\theta_a}(D^L)\cup f_{\theta_a}(D^U)$.
		
		\STATE Construct transition matrix $P_{\theta_{b}}(D)$ with $f_{\theta_b}(D)$ by Eq.~(\ref{graph}). 
		\STATE Construct transition matrix $P_{\theta_{a}}(D)$ with $f_{\theta_a}(D)$ by Eq.~(\ref{graph}).
		
		\STATE Initiate soft label set $Y_{\theta_a}(0) = Y_{\theta_b}(0) = [Y^L;$\textbf{0}$]$.
		
		\FOR { t = 1 to $T$}
		
		\STATE $Y_{\theta_{b}}(t) = P_{\theta_{b}}(D)*Y_{\theta_{b}}(t-1)$, \\
		$Y_{\theta_{a}}(t) = P_{\theta_{a}}(D)*Y_{\theta_{a}}(t-1)$.
		
		\STATE $Y_{\theta_{b}}^L(t) = Y_{\theta_{a}}^L(t) = Y^L$.
		
		\ENDFOR		
		
		\STATE $Y_{\theta_{b}}^U = Y_{\theta_{b}}^U(T)$, $Y_{\theta_{a}}^U = Y_{\theta_{a}}^U(T)$.
	\end{algorithmic}
\end{algorithm}

\subsection{Sample Selection via Transforming Features}
\label{32}
In the previous discussion, the change about local samples via feature transformation plays an important role to select reliable unlabeled samples. We attribute the change into two different variations. Firstly, as an off-the-shelf tool, label propagation algorithm is provided in Co-Space to assign a couple of soft labels for each sample. And the otherness between them, named as \textbf{neighborhood intrinsic variation}, is used to measure the change about intrinsic structure around each unlabeled sample. Secondly, we take a consideration in the labeled neighbors of unlabeled samples. The situation about local samples belong to each class are estimated in statistic, and its discrepancy in transformation is interpreted as \textbf{neighborhood category variation}. Finally, we incorporate both variations into an unified criterion to screen data.

\subsubsection{Neighborhood Intrinsic Variation}

In this subsection, neighborhood intrinsic variation will be formulated as the discussion below. We introduce the label propagation algorithm (LP), which is a key part of calculating the variation. As a wrapper algorithm, Eq.~(\ref{graph}) in Sect.~\ref{related} is utilized to construct the transition matrix $P$. Then an original
LP algorithm with $K$nn graph is demonstrated as Algorithm  \ref{Algorithm1}.

In the configuration we discuss, it needs some revision for Algorithm \ref{Algorithm1} to adapt to DCS. Firstly, since LP is a kind of fixed-feature transductive learning algorithm, $D$ in Algorithm \ref{Algorithm1} has been default as an extracted feature set for data. As for our framework, CNN extract features on the fly, and is also updated in the progressively training process. It motivates us to use $f_{\theta}(D)$ instead of $D$. Secondly, the algorithm is launched in Co-Space, which need two feature spaces to attain two label propagation outputs for comparison. Using <$\tilde{f}_{\theta_{b}}(D)$, $\tilde{f}_{\theta_{a}}(D)$> as input, the adaptive LP in Co-Space is shown in Algorithm  \ref{Algorithm2}.

As the illustration in Algorithm \ref{Algorithm2}, Co-Space leads to a couple of transition matrices <$P_{\theta_{b}}(D)$, $P_{\theta_{a}}(D)$> to predict soft label sets $Y_{\theta_{b}}^U$ and $Y_{\theta_{a}}^U$. Suppose normalized vector \textbf{y}$_{\theta_b}($\textbf{x}$)\in R^m$ belongs to $Y_{\theta_{b}}^U$ and normalized vector \textbf{y}$_{\theta_a}($\textbf{x}$)\in R^m$ belongs to $Y_{\theta_{a}}^U$. The non-consistency between \textbf{y}$_{\theta_b}($\textbf{x}$)$ and \textbf{y}$_{\theta_a}($\textbf{x}$)$, is performed as the neighborhood intrinsic variation in the transformation of sample \textbf{x}, composing the sample mining criterion about to mention. (in the iteration $t$, the $\theta_b$, $\theta_a$ refer to $\theta_{t-1}$, $\theta_t$). Since soft label is assigned through the intrinsic structure, which provided by an approximated manifold embedded in feature space\cite{Zhu2005Semi} according to $K$nn graph. It implies that in neighborhood intrinsic variation,  the change about local samples closer to \textbf{x} are more concerned.
  
%Category unlabeled neighbor variation represents the consistency of "smootheness"\cite{keylist}, namely, the intrinsic strucutre of training data after transformation. As for further discussion, we use $\begin{bmatrix} L_{ll}(\theta_a) & L^{T}_{ul}(\theta_a) \\ L_{ul}(\theta_a) &L_{uu}(\theta_a)  \end{bmatrix}$ and $\begin{bmatrix} L_{ll}(\theta_b) & L^{T}_{ul}(\theta_b) \\ L_{ul}(\theta_b) &L_{uu}(\theta_b)  \end{bmatrix}$ to denote transition matrics $P_{\theta_{b}}(D)$ and $P_{\theta_{a}}(D)$ respectively. In order to avianize the effect from labeled neighbor, we take two assumptions: 1). The labeled neighbors for \textbf{x} are close enough to determine the category of \textbf{x}. 2). The distances between \textbf{x} and its labeled neighbors, changes very little in feature space transformation. That means submatrices $L_{ul}(\theta_a)$ and  $L_{ul}(\theta_b)$ is treated as equal in transformation. Since
 
 %\begin{displaymath}
 %\begin{aligned}
 %Y^U &= (\mathbf{1}+L_{uu}+\cdots+L_{uu}^{T-1})*L_{ul}*Y^L \\ &\approx L_{uu}^{-1}*L_{ul}*Y^L
 %\end{aligned}
 %\end{displaymath}
  
%Under the assumption we mentioned, \textbf{x}'s soft label $\mathbf{y}(\mathbf{x})$ is mostly determined by submatrix $L_{uu}$. Hence,   the intrinsic structure based on the distribution of \textbf{x}'s neigborhood, can also be explored by the consistency of the soft label as well. Generally speaking, the category distribution variation focuses on the change about micro-structure around \textbf{x}.

\subsubsection{Neighborhood Category Variation}
\label{ncv} 
 Different from the intrinsic variation, neighborhood category variation focuses on the change of categories. In specific, it aims to search the classes with similarity changing apparently in statistic. Such similarity is estimated via measuring the local density variation of its labeled neighbors in feature transformation, which represents as a class-based matrix and calculated through a measurement deriving from Hellinger distance. For the sake of further discussion, we firstly introduce local labeled sample covariance matrix. 
 
 Specifically, we have a sample \textbf{x} and $f(\mathbf{x})$ is the corresponding feature.  $N(f(\mathbf{x}))$  the neighborhood around \textbf{x}, then the local sample covariance matrix for $f(\mathbf{x})$ is interpreted as follows:
 
 \begin{displaymath}
 \Sigma_{{f(\mathbf{x})}} = \frac{\underset{x^\prime \in N(f(\mathbf{x}))}{\sum}(x^\prime- \mu_{f(\mathbf{x})})^T (x^\prime- \mu_{f(\mathbf{x})})}{|N({f(\mathbf{x})})|-1}.
 \end{displaymath} 
 where $|N({f(\mathbf{x})})|$ denotes how many local samples in $N({f(\mathbf{x})})$, and $\mu_{f(\mathbf{x})} = \frac{\underset{x^\prime \in N({f(\mathbf{x})})}{\sum} x^\prime}{|N({f(\mathbf{x})})|}$ is the mean for local samples in $N(\mathbf{x})$. Considering different class belong to different distribution, we assume $\textbf{x}$ is classified as $y$. Then $f(\mathbf{x})$'s labeled neighbors belong to class $y$ denote as $N_y(f(\mathbf{x}))$ and the mean value in the neighborhood about class $y$ is rewrited to $\mu^y_f(\mathbf{x}) = \frac{kf(\mathbf{x})+\underset{\mathbf{x}^\prime \in N_y(f(\mathbf{x}))}{\sum} x^\prime}{|N_y(f(\mathbf{x}))| + k}$. The local labeled sample covariance matrix of class $y$ neighbors around $f(\mathbf{x})$ is defined as:
 
\begin{small}
	\begin{displaymath}
	\Sigma^y_{f(\mathbf{x})} = \frac{k(f(\mathbf{x}) - \mu^y_{\mathbf{x}})^T (f(\mathbf{x}) - \mu^y_{\mathbf{x}})+\underset{x^\prime \in N_y(f(\mathbf{x}))}{\sum}(x^\prime- \mu^y_{\mathbf{x}})^T (x^\prime- \mu^y_{\mathbf{x}})}{|N_y(f(\mathbf{x}))|+k-1}.
	\end{displaymath}
\end{small}
where $f(\mathbf{x})$ is arranged as one part of its labeled neighbors, and we treat it as class $y$ when covariance matrix $\Sigma^y_{f(\mathbf{x})}$ is considered. $k$ is a weight to balance the importance between $f(\mathbf{x})$ and its $y$ labeled neighbors. 

The covariance matrix $\Sigma^y_{{f(\mathbf{x})}}$ captures the local geometry density and statistic about labeled samples, which in the area around \textbf{x} and  belong to class $y$. After that, transformation distance is leveraged to measure the similarity between labeled neighbor in different classes. More specifically, for a sample \textbf{x} given class $y$, there is a Gaussian local distribution $p_{y}(f(\mathbf{x}))$ presenting as $\mathcal{N}(f(\mathbf{x}); 0, \Sigma^y_{f_(\mathbf{x})})$; then providing feature transformation $F_t($\textbf{x}$)$,  transformation distance deriving from Hellinger distance, is calculated as follows:

\begin{equation}
\label{HellingerDistance}
\begin{aligned}
\rho(F_t(\mathbf{x}),y; f_{\theta})
&\equiv 
 H(p_{y}(f_{\theta_{t-1}}(\mathbf{x})), p_{y}(f_{\theta_{t}}(\mathbf{x})) )
 \\
&= \sqrt{ 1 - \frac{ 2^{D/2}|\Sigma^y_{f_{\theta_{t-1}}(\mathbf{x})}|^{1/4} |\Sigma_{f_{\theta_(t)}(\mathbf{x})}^{y}|^{1/4}}{|\Sigma_{f_{\theta_{t-1}}(\mathbf{x})}^{y} + \Sigma_{f_{\theta_t}(\mathbf{x})}^{y}|^{1/2} } }
\end{aligned}
\end{equation}
where $D$ is the dimensionality of the feature space, and the $|\Sigma^y_{f(\mathbf{x})}|$ means the determinant of matrix $\Sigma^y_{f(\mathbf{x})}$.

	\begin{algorithm}[htb]
		\caption{ Neighborhood Category Variation}
		\label{Algorithm3}
		\begin{algorithmic}[1]
			
			\REQUIRE ~~\\
			
			Unlabeled data $D^U$, Co-Space <$f_{\theta_{t-1}}(D)$, $f_{\theta_t}(D)$>, $s$.\\
			
			\ENSURE ~~\\
			Transformation matrix set $\{M_{f_{\theta}}(F_{t}(\mathbf{x}))| \mathbf{x}\in D^U\}$.
			
			\FOR { i = 1 until $|D^U|$}
			
			\STATE Obtain $\mathcal{Y}(F_t(\mathbf{x}_i),s)$, where $\mathbf{x}_i\in D^U$ and $F_t(\mathbf{x}_i)\in $ <$f_{\theta_{t-1}}(D)$, $f_{\theta_t}(D)$>.
			\FOR { k = 1 until $s$}
			
			\STATE For $y_k\in \mathcal{Y}(F_t(\mathbf{x}_i),s)$, obtain $\Sigma^{y_k}_{f_{\theta_{t-1}}(\mathbf{x}_i)}$ and $\Sigma^{y_k}_{f_{\theta_{t}}(\mathbf{x}_i)}$.
			
			\STATE Obtain $\rho(F_t(\mathbf{x}_i),y_k; f_{\theta})$ via Eq.~(\ref{HellingerDistance}) from step 4.		 
			
			\ENDFOR
			\STATE Obtain $\{\kappa(F_{t}(\mathbf{x}_i),y;f_{\theta})| y\in \mathbf{Y}\}$ via Eq.~(\ref{kappa}).
			
			\STATE Obtain $M_{f_{\theta}}(F_{t}(\mathbf{x}_i))$.
			\ENDFOR
		\end{algorithmic}
	\end{algorithm}

In the formulation above, a major problem comes from the computational complexity, which increasing in pace with the size of class number $m$. But thanks to the locality, we are just interested in an area around \textbf{x} and unnecessary to take all classes into account. 
In specific for image \textbf{x}, we choose the intersection of its labeled neighborhoods before/after transformation. The major top-$s$ categories in the intersection are considered by Eq.~(\ref{HellingerDistance}), and the top-$s$ category set for image \textbf{x} denotes as:

%Let $x$ be a local labeled sample around \textbf{x} and labeled as $y$, then, $x$ belong to a preserved labeled neighborhood $\mathcal{U}_y(F_{t}(\mathbf{x}))$, is described in the feature transformation setting of \textbf{x} as follows:

	%\begin{displaymath}
	%\begin{aligned}
	%x\in\mathcal{U}_y(F_{t}(\mathbf{x})) \iff & \ f_{\theta_{t-1}}(x)\in N_y(f_{\theta_{t-1}}(\mathbf{x}))) \\ & \cap  \ f_{\theta_t}(x)\in N_y(f_{\theta_t}(\mathbf{x}))\\
	%\end{aligned}	
	%\end{displaymath}

	\begin{displaymath}
		\mathcal{Y}(F_t(\mathbf{x}),s) \subset \mathbf{Y}
	\end{displaymath}

Heuristically in implementation, we only choose a few classes ($s$ less than 5, case by case) as the consideration in labeled neighborhood, then for other categories, the related Hellinger distances are set as infinity. We use $h$ to denote exponential decay function $exp(-x)$ in DCS, then for a specific class $y$ , Eq.~(\ref{HellingerDistance}) is reformulated to $\kappa$ as:

\begin{small}
	\begin{equation}
	\label{kappa}
	\kappa(F_{t}(\mathbf{x}),y;f_{\theta}) \!= \! \left\{
	\begin{aligned}
	& \ \frac{h(\rho(F_t(\mathbf{x}),y; f_{\theta_t}))}{\underset{y'\!\in\! \mathcal{Y}(F_t(\mathbf{x}),s)}{\sum} h(\rho(F_t(\mathbf{x}),y'; f_{\theta}))} & y\!\in\! \mathcal{Y}(F_t(\mathbf{x}),s)\\
	& \ 0 & otherwise\\
	\end{aligned}
	\right .
	\end{equation}
\end{small}

\subsubsection{Sample Mining Criterion}
In transformation about \textbf{x}, the neighbors variation is estimated from two aforementioned points of view. Further, we assemble both variations into one criterion. We have a feature transformation matrix deriving from Eq.~(\ref{kappa}) as:

\begin{equation}
\label{MM}
M_{f_{\theta}}(F_{t}(\mathbf{x})) = \begin{bmatrix}\begin{smallmatrix}
\mathbf{\kappa}(F_t(\mathbf{x}),1; f_{\theta}) &  0  & \cdots\ &0\\
0  &  \mathbf{\kappa}(F_t(\mathbf{x}),2; f_{\theta})  & \cdots\ & 0\\
\vdots   & \vdots & \ddots  & \vdots  \\
0 & 0  & \cdots\ & \mathbf{\kappa}(F_t(\mathbf{x}),m; f_{\theta})\\
\end{smallmatrix}
\end{bmatrix}
\end{equation}
where each row (column) in the matrix refers to a specific class in \textbf{Y}, and the workflow shows in Algorithm \ref{Algorithm3}.

Appending the result in Algorithm \ref{Algorithm2}, we obtain the confidence function $R(\mathbf{x};\theta_t)$ below, scoring the reliability for each unlabeled image \textbf{x}:

\begin{equation}
\label{1}
\begin{aligned}
R(\mathbf{x};\theta_t) &= r_b(\mathbf{x};\theta_t)^Tr_a(\mathbf{x};\theta_t)\\ 
s.t. \ &\mathbf{x} \in D_t^U \\
 r_b(\mathbf{x};\theta_t) &= \frac{\sqrt{M_{f_{\theta}}(F_{t}(\mathbf{x}))}\mathbf{y}_{\theta_{t-1}}(\mathbf{x})}{|\sqrt{M_{f_{\theta}}(F_{t}(\mathbf{x}))}\mathbf{y}_{\theta_{t-1}}(\mathbf{x})|}; \\
 r_a(\mathbf{x};\theta_t) &= \frac{\sqrt{M_{f_{\theta}}(F_{t}(\mathbf{x}))}\mathbf{y}_{\theta_{t}}(\mathbf{x})}{|\sqrt{M_{f_{\theta}}(F_{t}(\mathbf{x}))}\mathbf{y}_{\theta_{t}}(\mathbf{x})|};  
\end{aligned}
\end{equation}
which implies the cosine similarity between vectors $r_b(\mathbf{x};\theta_t)$ and $r_b(\mathbf{x};\theta_t)$.
 Those samples with high score in Eq.~(\ref{1}) will be chosen, and get annotation by the rule as:
 
 \begin{equation}
 \label{2}
 \begin{aligned}
  L(\mathbf{x};\theta_t) &= \underset{y}{\mathbf{argmax}}\{v(\mathbf{x};\theta_t)\}\\
  s.t. \ v(\mathbf{x};\theta_t) &= r_b(\mathbf{x};\theta_t)\circ r_a(\mathbf{x};\theta_t)
 \end{aligned}
 \end{equation}

 where $\circ$ is the Hadamard product for matrices (vectors) with same dimensions, $\mathbf{argmax}\{v\}$ choose the biggest scalar value from the entries of vector $v$ .
 
 Function $R$ is used to measure the consistency in unlabeled samples via feature transformation, in which each class in label set is considered and contributes to the confidence score in Eq.~(\ref{1}). If the value is more than a pre-defined threshold, the unlabeled sample will be selected and labeled as a class with largest contribution in Eq.~(\ref{1}).
 
 There are $M$ samples chosen for each iteration at most, and threshold $th$ promise confidence score always bigger than a constant. The candidates with pseudo labels are used to enrich the labeled data pool and the cycle repeats as the progressive process in DCS.

\subsection{Further Discussion about DCS} 
\label{33}
 We discuss how the consistency is estimated through the criterion. As a careful observation in Eq.~(\ref{1}), reliability function $R$ calculates the cosine distance between a couple of class re-weighted label propagation results in Co-Space, and the rebalanced weights for each classes, is provided by feature transformation matrix $M_{f_{\theta}}(F_{t}(\mathbf{x}))$. Our intuition comes from the noninformative problem about LP\cite{nadler2009semi}. Regardless of feature transformation matrix, Eq.~(\ref{1}) degenerates to a simple cosine similarity between the couple of soft label in Co-Space, showing extremely unstable results in incrementally features transforming setting in our experiment. This is explained by the continuous augmentation of large scale training data, which triggering the noninformative problem. Feature transformation matrix helps the sample mining strategy focus on a few classes mainly acting on a provided sample, and restrain the redundant class information propagated through unstable relationship structure, which constructed from unlabeled samples represented by immature features. 
 
\begin{algorithm}[htb]
	\caption{ Deep Co-Space (DCS) Sample Mining  }
	\label{Algorithm4}
	\begin{algorithmic}[1]
		
		\REQUIRE ~~\\
		
		Labeled image dataset $D^L$ and unlabeled image dataset $D^U$; The label set $Y^L$ corresponding to dataset $D^L$; CNN-based model $f$ with a pre-trained parameter $\theta^o$, pre-setted max iteration M.\\
		
		\ENSURE ~~\\
		Well-trained CNN-based model $f_{\theta^*}$;
		pesudo-labeled image set $D^*$.
		
		\STATE {Initiate $f$ by $\theta^o$ and obtain $f_{\theta_0}$; $D^L_0 = D^L$ and $Y^L_0 = Y^L$; $D^U_0 = D^U$; $D_0 = \{D^L_0, Y^L\} \cup D^U_0$; $D^* = \emptyset$; $\{D_s,Y_s\} = \emptyset$}.
		
		\STATE Obtain $\theta_1$ via fine-tuning $f_{\theta_{0}}$ with $D_{0}$.
		
		\FOR { t = 1 until M}
		
		\STATE Obtain Co-Space <$f_{\theta_{t-1}}(D)$, $f_{\theta_t}(D)$> through feature extraction by $f_{\theta_{t-1}}$ and $f_{\theta_{t}}$.
		
		\STATE Utilize LargeVis\cite{tang2016visualizing} to  <$f_{\theta_{t-1}}(D)$, $f_{\theta_t}(D)$> ,  Obtain Co-Space <$\tilde{f}_{\theta_{t-1}}(D)$, $\tilde{f}_{\theta_{t}}(D)$>.
		
		\STATE Runs Algorithm \ref{Algorithm2} to obtain soft label sets $Y^U_{\theta_{t-1}}$ and $Y^U_{\theta_{t}}$ for $D^U_t$.
		
		\STATE Runs Algorithm \ref{Algorithm3} to obtain feature transformation matrix set for $D^U_t$.
		%\STATE Obtain feature transformation matrix set for <$\tilde{f}_{\theta_{t-1}}(D)$, $\tilde{f}_{\theta_{t}}(D)$> by Eq.\ref{MM}
		
		\STATE Use Eq.~(\ref{1}) to select the top $M$ samples with highest scores and annotate them in the principle of Eq.~(\ref{2}).
		
		\STATE  {$D^L_{t}=D^L_{t-1}\cup D_s$,
			$Y^L_{t}=Y_{t-1}\cup Y_s$, 
			$D^U_t=D^U_{t-1}/D_s$, 
			$D_t = \{D^L_{t}, Y^L_{t}\} \cup D^U_t$}.
		\STATE Fine-tune CNN $f_{\theta_{t-1}}$ with $D_t$, obtain $\theta_{t}$; $D^* = D_s\cup D^*$.
		
		\ENDFOR
	\end{algorithmic}
\end{algorithm} 
 
 In another point of view, we explain the strategy as measuring "micro-structure and "macro-structure around data. In "micro-structure about data in semi-supervised learning, data present as a manifold where their classes change smoothly\cite{zhu2003semi}. Intrinsic structure about data presenting by $k$nn, captures the local property around the considered data instance \textbf{x}. In case of that, we use a couple of soft labels to contrast the category change across intrinsic structures in different spaces. Differently, "macro-structure assuming data belong to same class should cluster together\cite{chapelle2005semi}. We use local labeled sample covariance of each class to perceive the geometric property change around \textbf{x} (density, shape and so on), and the class with steady geometric property is more preferable.

 %DCS depends on deep neural network but does not rely on neural network architecture. Hence it could also be treated as a generalized semi-supervised deep learning framework, and readily grafted to deep supervised learning (DSL) model while just few data with label have been given.
 
%% -----------------------------------------------------  Obtain the Labels

 In the implementation of DCS, $f_{\theta}(D)$ as CNN-based features, are inappropriate as a direct input to calculate transition matrix $P_{\theta}(D)$. Due to in visual classification, $f_{\theta}(D)$ often coming from fully connected layer, the extracted feature is high-dimensional. Referring to the analysis in \cite{nadler2009semi}, GSSL algorithm applied in high-dimensional scenario inevitably runs into a common problem, leading to the value of label function
 for unlabeled sample constant almost everywhere in feature space. An alternatives to compromise the problem is dimensionality reduction. However, linear reduction methods\cite{karnin2015online} decompose the local correlation among samples, and when we choose to preserve the locality\cite{maaten2008visualizing}, the computational complexity will be demanding. In the up-to-date related researches,  LargeVis\cite{tang2016visualizing} is an ideal option in balance. As an innovative approach to make data visualization, LargeVis can also be treated as a locality preserving technique for dimensionality reduction. The computational complexity of LargeVis is $\mathcal{O}(lMn)$ ($n$ is the number of images we are about to cope with; $M$ and $l$ are the dimensionality of the original space and target space respectively.), keeping sample mining strategy  computationally
 feasible in the incremental processing setting.

Then we discuss the computational complexity about DCS. Since DCS is an incremental learning framework to gradually process large scale data, we only observe one iteration in the cycle. The LargeVis achieves dimensionality reduction with complexity of $\mathcal{O}(lMn)$, which linearly related to sample size $n$. Using LP algorithm within Co-Space in the generic style, the graph construction and propagation have a total time cost as $\mathcal{O}(2n^2)$. Afterwards, in order to attain feature transformation matrix $M$, we 
runs Algorithm \ref{Algorithm3}. It seems complicated but the total computation cost is $\mathcal{O}(ns(N+k))$, where $N$ is the maximal number of labeled neighbors for each unlabeled sample and $k$ is the balancing weight in the calculation of $\Sigma^y_{f(\mathbf{x})}$. As the discussion regardless of considering feature extraction and the fine-tuning model, the bottleneck in complexity is the graph construction for LP. Actually, there exists a more scalable alternative to build relational graph\cite{liu2010large} in linear $n$ time complexity. Moreover, inversely to the progressively data processing setting, we also sift out unlabeled samples selected previously, which improving the speed to account relationship between features in our DSSL experiment.

The work flow of DCS thoroughly shows in Algorithm \ref{Algorithm4}

%the determinant $\Sigma^y_{f(x)}$, which calculated through Co-Space <$f_{\theta_{t-1}}(D)$, $f_{\theta_{t}}(D)$>, is always fixed at $0$. It is because $\Sigma^y_{f(x)}$ comes from a statistical estimation on a bunch of features lying in a high-dimensional feature space, thus, the local sample magnitude is far less than the dimension. Following the routine in subsection 1), we use <$\tilde{f}_{\theta_{t-1}}(D)$, $\tilde{f}_{\theta_{t}}(D)$> as a substitute to calculate $\rho(F_t(\mathbf{x}),y; f_{\theta})$ and this problem is solved. Another point which shouldn't be forgotten is the computational complexity. The series of covariance matrics $\{\Sigma^y_{f(\mathbf{x})}|$y  $\in$\textbf{y}$\}$ capturing all classes behaviors around \textbf{x}, leads to

%----------------------------------------------------------------------------------- 
% --------------------------------------------------------------------------------------------------------------------------------------------------------------------------------------------------------------

%%%%%%%%%%%%%%%%%%%%%%%%%%%%%%%%%%%%%%%%%%%%%%%%%%%%%%%%%%%%%%%%%%%%%%%%%%%
%%%%%%%%%%%%%%%%%%%%%%%%%%%%%%%%%%  Experiments. %%%%%%%%%%%%%%%%%%%%%%%%%%%%%%%%%
%%%%%%%%%%%%%%%%%%%%%%%%%%%%%%%%%%%%%%%%%%%%%%%%%%%%%%%%%%%%%%%%%%%%%%%%%%%
\begin{figure}[t]	
	\includegraphics[width=3.3in]{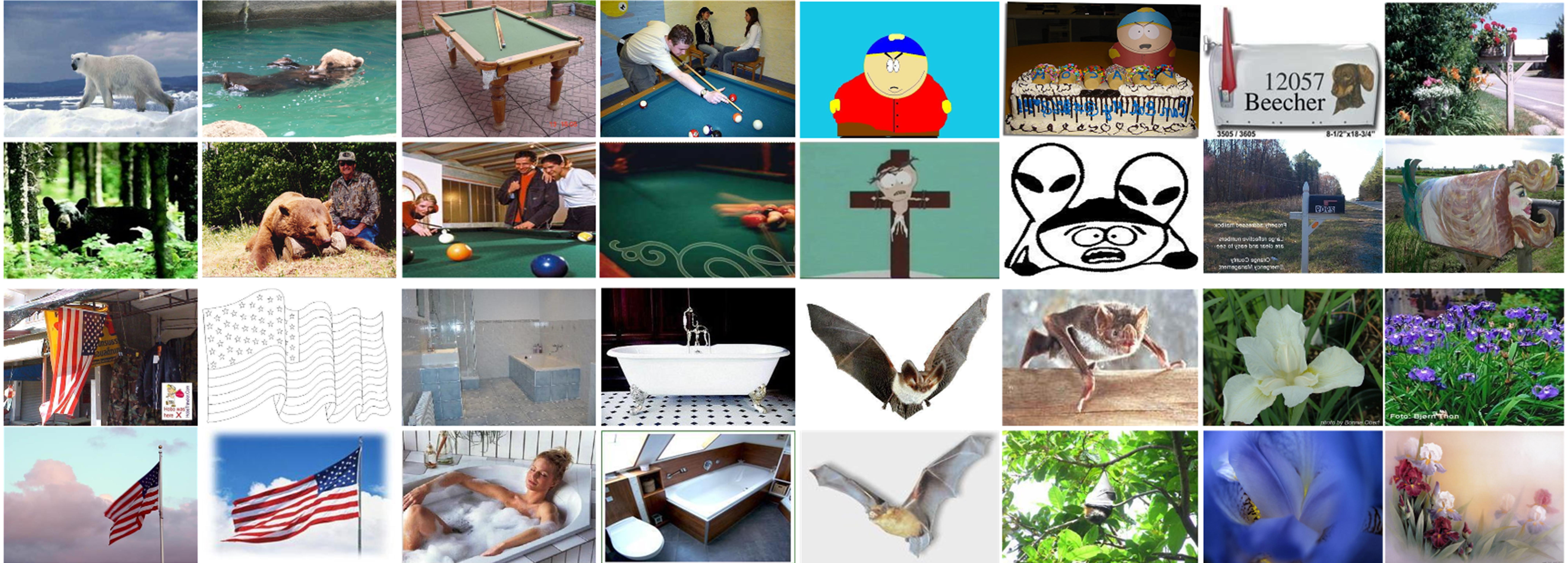}
	\caption{Samples of images about Caltech-256. We selected 22,746/6,612 of RGBs for training and testing respectively. }
	\label{fig4}
\end{figure}

\begin{figure}[t]	
	\includegraphics[width=3.3in]{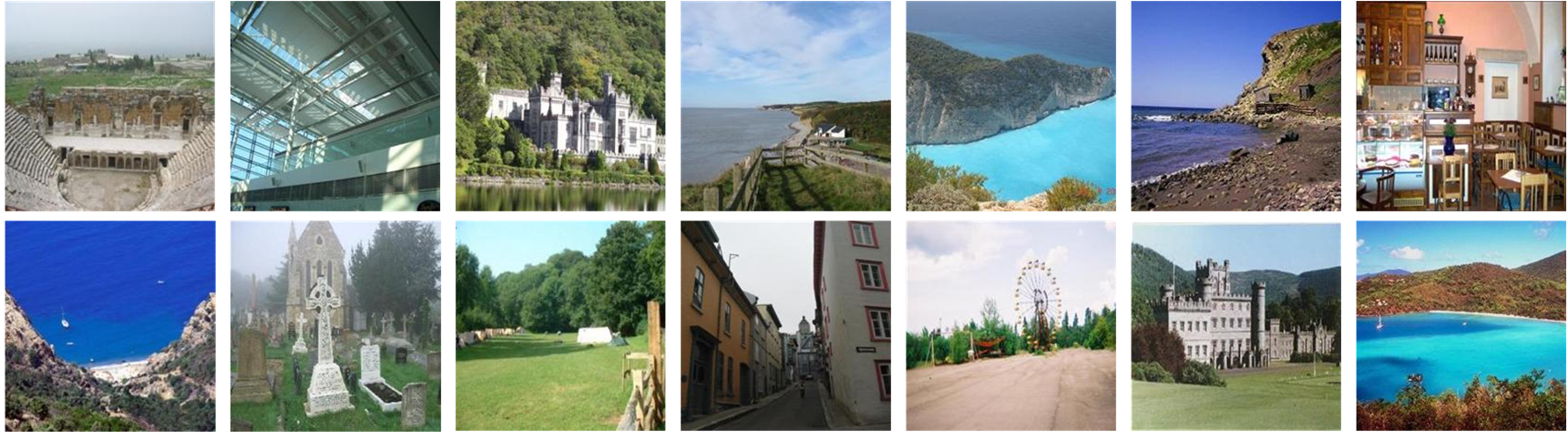}
	\caption{Samples of images about SUN-397. We chose a subset of them in the experiment setting.}
	\label{fig3}
\end{figure}

\begin{figure}[t]	
	\includegraphics[width=3.3in]{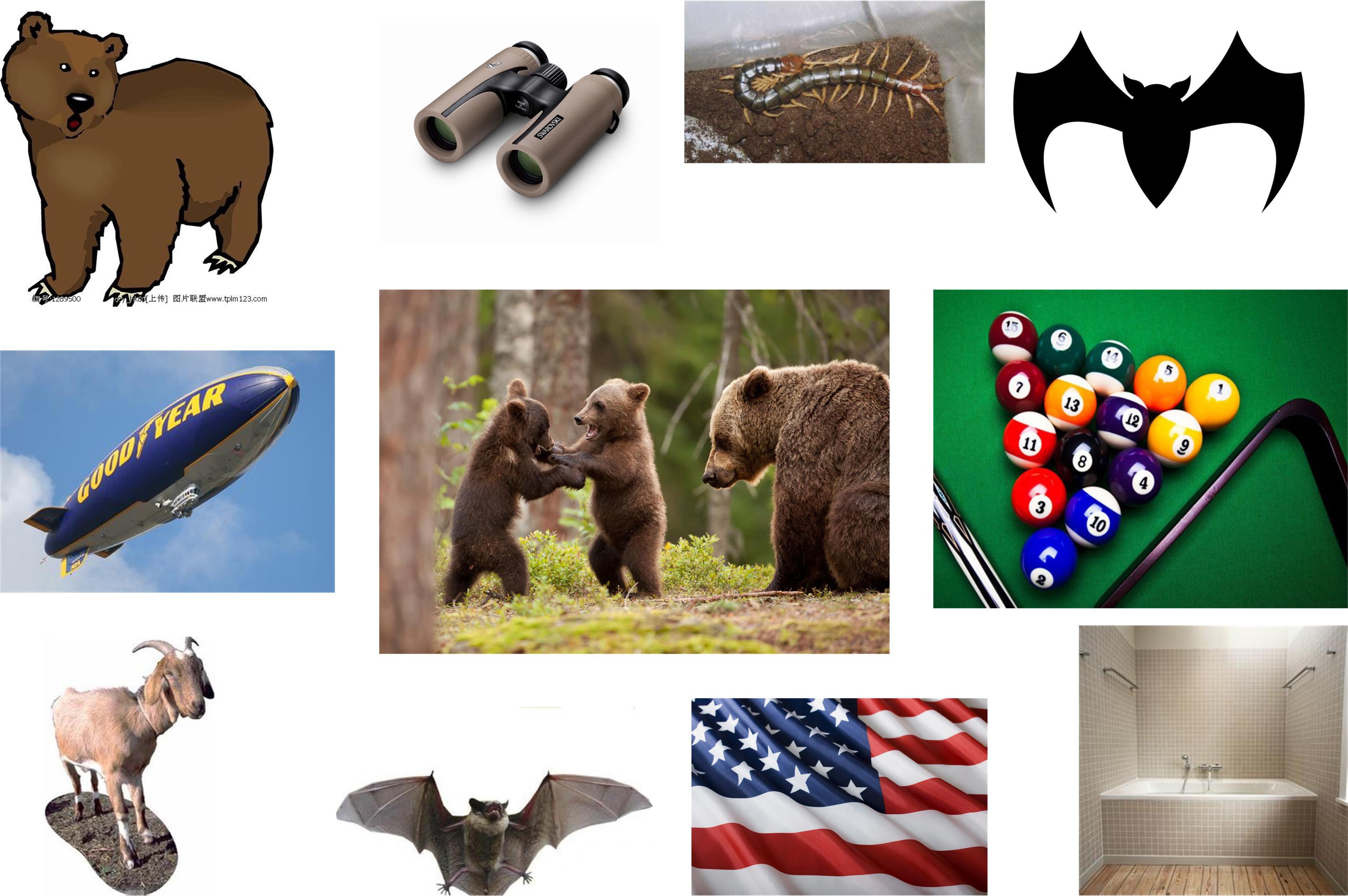}
	\caption{Samples of web images collected by the search engine. They are full of noise and we treat them as unlabeled images to expand the Caltech-256.}
	\label{fig5}
\end{figure}

\section{Experiments}

\label{exper}

We evaluate our DCS in two different semi-supervised learning settings to validate its effectiveness in Sect.~\ref{sec1}. Moreover, we further analyze the components of our DCS to clarify their contributions in Sect.~\ref{sec2}.

\subsection{Empirical Study}

\label{sec1}
\begin{table*}
	\begin{center}
		\caption{Comparison of our results with several comparison network architectures on Caltech-256-VGG (DAE-PL and SWWAE have been augmented with web images)}
		\label{t1}
		\begin{tabular}{|c||c|c|c|c|c|c|c|c|c|c|c|c|ccr}
			\hline
			Architectures in DSSL							
			
			&\multicolumn{3}{|c|}{VGG (fully-supervised learning)}  & \multicolumn{4}{|c|}{{VGG-based DAE-PL}}  & \multicolumn{4}{|c|}{VGG-based SWWAE}   	\\
			\hline 	
			\hline
			
			Percentage of labeled data &{9.4\%}&{18.8\%}&100\%
			&\multicolumn{2}{|c|}{{9.4\%}}
			&\multicolumn{2}{|c|}{{18.8\%}}
			&\multicolumn{2}{|c|}{{9.4\%}}
	        &\multicolumn{2}{|c|}{18.8\%}
	        \\
	        \hline
	        DCS involved &No &No &No &No &Yes &No &Yes &No &Yes &No &Yes \\
	        \hline
	        Error rates								 	&{Overfitting} & {45.6\%} &23.47\%								& {42.26}\%		&{\textbf{41.57}}\% 	 &{37.20\%} &{\textbf{36.45}}\%	&{31.52\%} 	 								& {\textbf{30.05}}\% 								& 29.42\%		&\textbf{26.73}\%
	        \\ 
	        \hline
	        \hline

		\end{tabular}
	\end{center}
\end{table*}

\textit{Experimental Setting} 
Thanks to be independent of any specific network architecture, our DCS can be easily grafted to many different deep convolution-based models, and improve their performances in visual classification aided by large scale unlabeled images. To justify this, we conduct two experiments to evaluate DCS with two branches of deep learning models. The first one is \emph{Deep Semi-Supervised Learning} model (\textbf{DSSL}), in which the convolutional network architecture is invented to receive labeled images and unlabeled images in parallel. while the second one is \emph{Standard Supervised Neural Network} model (\textbf{SSNN}), namely, the network must be trained under full supervision. Accordingly, we utilize labeled information to initiate SSNN models, then fine-tune them with an augmented labeled image set in a progressive style. Then in each iteration of DCS, the labeled image pool will be enriched by reliable pseudo-labeled samples, leading to an updated CNN for next iteration. We set the upper limit of labeled data augmentation as 1000 and the initial base learning rate as 0.001, then, all CNNs are updated via stochastic gradient decent algorithm with the momentum 0.9 and weight decay 0.0005. After the dimensionality reduction by LargeVis, the dimension in Co-Space is reduced to 15 before label propagation.

%In specific, we leverage stacked what-where auto-encoder (SWWAE)~\cite{zhao2015stacked} as the architecture in implementaion; then, train SWWAE with all images in each iteration, update label information in image database and proceed to loop.

%The couple of models <$f_{\theta_{t-1}}, f_{\theta_{t}}$> are treated as deep feature extractors and leads to Co-Space <$f_{\theta_{t-1}}(D), f_{\theta_t}(D)$>, where $D$ is the target training set and $f_{\theta_t}$ denotes the convolution network in iteration $t$ ($f_{\theta_0}$ means pre-trained model). For all networks, we extract features from fc-7 layer with 4096 in dimension. 

Since Co-Spaces are sequentially constructed in DCS, it requires the scalability about the LP algorithm in DCS. In our empirical experiment, $\delta$ is set $0.9$ and the algorithm runs 50 times for each iteration. Besides, in $k$ nearest labeled neighbors setting in Eq.~(\ref{HellingerDistance}), the closest $300$ local labeled samples are considered and top 5 classes with most samples number will be selected to compute class-specific Eq.~(\ref{2}).

%There are two reasons make us choose the two visual recognition datasets instead of standard dataset like face recognition dataset CACD, SVHN,e.g., which have been frequently adopted in SSL. Firstly, the DCS aims to detect the local stable structure in data space through 

%The max iteration is set $6$ to all experiment, and $1000$ unlabeled samples would be selected in each iteration.    

\subsubsection{\textbf{The Experiment of DSSL}} The experiment is conducted on public object recognition benchmark Caltech-256~\cite{Griffin2007Caltech}, which includes 30,607 images in total (Please see Fig.~\ref{fig4} for more details). We randomly select 80\% images of each class as training data and the rest 20\% images are treated as testing data, then there are 29,358 RGB images, which contain 22,746/6,612 for training and testing respectively.

Since the training data is insufficient in DSSL experiment, we collect additional unlabeled images to address this issue. Specifically, we utilized python-tools web-crawler to collect images based on the keywords as all categories in Caltech-256, and images of which size less than 50*70 or larger than 500*700 are screened out (Please see Fig.~\ref{fig5} for more details). We select 100 candidates as unlabeled images for each class in the rest, and there are 25,600 images in total to expand the original Caltech-256. As for semi-supervised training, different percentage of samples in original Caltech-256 are selected as an initial labeled images pool, and the rest and web-crawled images are treated as unlabeled data in the progressive learning framework.

\textit{Implementation Details}: Deriving from VGG architecture, two auto-encoder nets are employed to evaluate the efficacy of DCS. The first is a conventional    denoising auto-encoder trained in the principle of Pseudo-Label (DAE-PL)~\cite{lee2013pseudo}; the second model is stacked what-where auto-encoder (SWWAE)~\cite{zhao2015stacked} with all skip-connections activated. Both of them are well-known DSSL models based on reconstruction. Considering the mirror architecture of the both models, we take 16-layer VGG as the encoding pathway and initiate decoding layers with Gaussian random noise, then make deep unsupervised learning to pre-train whole the architecture with ImageNet ILSVRC 2012 dataset. In the pre-training and fine-tuning phase of DCS, spatial batch normalization\cite{ioffe2015batch} layers are leveraged to enhance the network performance for faster convergence.

\textit{Comparison and Analysis}:  In order to testify whether DCS is competent to boost the performance of DSSL models, we take both aforementioned models as the comparisons to their DCS-armed versions. Further, we validate the four DSSL models in the ratio of 9.4\% (80\% unlabeled images and web-images without label) and 18.8\% (60\% unlabeled images and web-images without label) respectively. Due to the class imbalance in Caltech256, both of the ratio settings have already been extream situations (The amount of labeled images in some class is less than 10). The ratio setting is able to prevent the minority class from collapse in performance and keeps the task challenging to demonstrate the efficacy of DCS. Besides, we introduce the original VGG as a baseline model in supervised learning, which shares the same encoding pathway in DAE-PL and SWWAE for comparison. Although it is not a DSSL model, the VGG is still able to show how much a fully-supervised model can be improved in total by progressively augmenting unlabeled image samples. As a comparison of the fully supervised learning, the supervised VGG is trained with 100\%, 9.4\%, 18.8\% labeled samples in Caltech-256 without web images augmentation respectively.

Table~\ref{t1} illustrates the results based on error rates. As we can see, DSSL models (DAE-PL and SWWAE) equiped with DCS outperform their original versions; and SWWAE+DCS trained with 18.8\% labeled images shows close performance to the VGG in the setting of fully-supervised learning with 100\% labeled images. This demonstrates that the performance of deep semi-supervised neural network can be enhanced by our DCS with unlabeled data. In further discussion, we note that unlabeled images from the Internet are often full with intra-class variation of the visual appearance, which tends to bring mild negative effects to the original SWWAE, which employs auto-encoder to reconstruct all unlabeled images to learn a latent expression. Besides, benefiting from the proposed sample mining criterion for reconstruction, SWWAE-based DCS shows the resistance of intra-class variation, and also reassuringly brings about more category information to obtain a clear performance gain.

\subsubsection{\textbf{The Experiment of SSNN}} Under this experiment setting, we evaluate DCS on two public visual classification databases: SUN-397~\cite{Xiao2014SUN} and the original Caltech-256 ~\cite{Griffin2007Caltech} (no web image augmentation). SUN-397 is a large scale of images for scene categorization, whose image number across categories varies from 100 to over 2000. SUN-397 contains 397 classes and 108,754 images in total (Please see Fig.~\ref{fig3} for more details). We use the subset of SUN-397, including top 80 classes in the number for the experiment respectively. Both datasets are split as training set and test set with a ratio 4:1 in all the following SSNN experiments. 

To the contrary of DSSL experiment where the networks are initiated with all images, the network in the SSNN experiment starts with only a few labeled images. This leads to the overfitting when the ratio of initial labeled images is quite small (e.g., the fully-supervised learning with 9.4\% labeled images in previous SSNN experiment). To address this issue, several tricks are leveraged. Specifically, when labeled images are too few to support the supervised pre-training from overfitting, we utilize the auto-encoder as the unsupervised pre-training step, then for the minority classes whose number of labeled images are less than 20, we evenly oversample them to balance both the pre-training and fine-tuning in DCS. In those minority classes, images are assigned labels by the TSVM~\cite{joachims1999transductive} trained with the extracted features at each step in DCS. Those heuristic methods help relieve the initiation problem that occurs in the SSNN experiment, however in practice, the SWWAE or DAE-PL are the more preferable architectures to deal with the case of less labeled images for the implementaion of DCS.

We account for 40\%, 45\% and 50\% for the original Caltech256 as initial labeled samples (We have shown the case with less labeled images within Caltech256 in the Table~\ref{t1}), and  20\%, 30\%, 40\%, 45\% and 50\% for the SUN397 database. When labeled images less than 30\% in the SUN397 experiment, we introduce the previous tricks to prevent the DCS training from overfitting.

%In term of dataset Caltech-256, we adopt the reorganization in the DSSL experiment except for web-image augmentation. 

%The results on the two datasets are reported in Table.2 and Table.3, the accuracy has been used as the metric of evaluation.

\textit{Implementation Details}: As for the network architecture, Vanilla Alexnet is implemented in the experiment about Caltech-256 without web images expansion, while VGG is applied in the subset of SUN-397. The corresponding parameters obtained on ImageNet ILSVRC 2012 dataset are used to initiate these two networks.

\textit{Comparison and Analysis}: We compare our DCS with other incrementally SSL training frameworks: i) Yarowsky algorithm (YA)~\cite{Haffari2012Analysis}. On the purpose of a significant comparison with this methods, we utilize deep learning architecture in YA same as DCS; ii) Deep learning via semi-supervised embedding (DSSE)~\cite{weston2012deep}. DSSE tends to build the deep network with the relationship between samples. It is regrettable that no existing relation information is provided except for partially labeled data. Therefore, we make the modification of the DSSE and adapt the algorithm to the incremental learning experiment setting. Specifically, each couple of images with same label is viewed as a close relationship; then some unlabeled samples with high confidence (small entropy loss) at each iteration are assigned a corresponding label. The modification promises all unlabeled samples without relationship given are taken into consideration for training. Similarly, we use the convnet with the same configuration as DCS. 3). Adaptive semi-supervised learning (ASL)~\cite{Wang2014Large}. ASL is not a deep learning algorithm, yet still keeps well-performed in some visual recognition benchmarks. We take it as a conventional feature-fixed method for comparison, and let an aforementioned CNN models to extract features as input. Finally, labeled samples have been used to train a baseline model, which named "labeled-data-CNN" in the tables.  

\begin{figure}[t]

	\center

	\includegraphics[width= 0.9 \columnwidth]{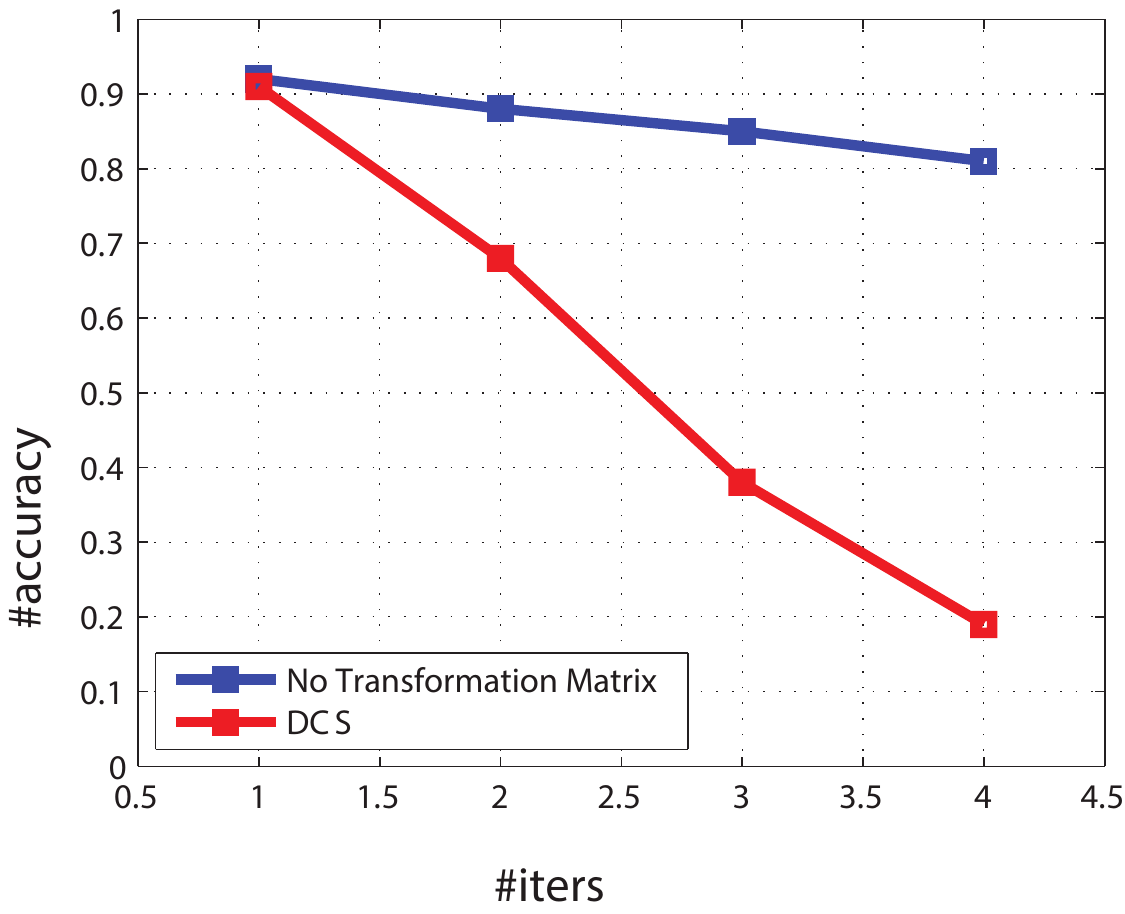}
	
	\caption{The diagram above demonstrates the result of component analysis about the results with/without transformation matrix. Axis x and y denote how many times of the iteration and the corresponding label prediction accuracy of the selected unlabeled images. Red/Blue line comes from DCS/Eq.~(\ref{3}) in the DSSL experiment}
	
	\label{ca}
	
\end{figure}

%It is obvious that whatever image database in the experiment, more labeled images for the database initialization, leads to a better performances under the same network architecture. 

In the experiment of SUN-397 dataset, the results are shown in Table \ref{t3}. As we can see, when initial labeled images for each class is sufficient, DCS out-performs all the compared methods. This justifies the effectiveness of the our DCS. However, according to the results for the experiment with Caltech256 in Table \ref{t2} and the 20\% situation of SUN397 in Table \ref{t3}, DCS is less prominent in the comparison with the other methods. We explain the result in two reasons. Firstly, Caltech-256 shares some categories with ImageNet ILSVRC 2012 which have been used to pre-trained the deep model in the supervised learning style. It makes the performance of those classes stuck in bottleneck and constrain the whole dataset performance. Secondly, according to Eq.~(\ref{HellingerDistance}), each unlabeled sample has a feature transformation matrix based on its labeled neighbors. Small proportion of each class leads to less labeled neighbors for each unlabeled sample, and increases the variance in the calculation of transformation matrix. The setting with 50\% labeled data expand the labeled neighbors, which relieve the problem and helps the model achieve a performance better than the other algorithms.

\subsection{Component Analysis}

\label{sec2}

The feature transformation matrix is a core in our DCS framework. For further demonstration of its contribution, we revise Eq.~(\ref{1}) and Eq.~(\ref{2}), and design a relevant component analysis with a new sample selection criterion as below:

\begin{table}[tb]
	
	\begin{center}
		
		% \footnotesize
		
		\caption{Comparison of our results with several comparison methods on SUN-397-80-VGG}
		
		\label{t3}
		
		\begin{tabular}{|l||c|c|c|c|c|c|cccccr}
			
			\hline
			
			&{20\%}  &{30\%}  &40\% &45\% &50\%  	   	\\

			\hline

			labeled-data-CNN		  													&{38.45\%} 		&{49.78\%}							&60.65\% 	  													&67.89\% 		&73.07\%					\\ 		
			
			ASL
			
			\cite{Wang2014Large}   													&{33.94\% }		&{45.46\%} &59.95\% 	    			          							& 		66.62\%		& 68.47\%		\\

			YA \cite{Haffari2012Analysis} 	  													&{24.76\% }		&{33.70\%}
			&47.77\% 				& 		57.80\%		&64.80\%	 \\

			DSSE \cite{weston2012deep}  													&{\textbf{41.65}\% }		&{46.84\%}
			&51.24\% 	    			          							& 		57.48\%		&67.97\%		\\
			
			\hline
			
			DCS  			  													&{38.96\% 	}	&{\textbf{52.07}\%}						&\textbf{61.78}\% 						     							&\textbf{69.66}\%			&\textbf{74.56}\%			\\
			
			\hline
			
			\hline
			
		\end{tabular}
		
	\end{center}
	
\end{table}

\begin{table}[tb]
	
	\begin{center}
		
		% \footnotesize
		
		\caption{Comparison of our results with several comparison methods on Caltech-256-Alexnet (no augmentation with web images)}
		
		\label{t2}
		
		\begin{tabular}{|l||c|c|c|ccccccr}
			
			\hline
			
			\hline 										
			
			& 40\%  & 45\%  & 50\% 	   	\\
			\hline

			\hline

			labeled-data-CNN								&60.05\% 	  													&63.87\% 		& 68.98\%					\\

			ASL
			
			\cite{Wang2014Large}&\textbf{60.96}\% 	    			          							& 		62.22\%		& 65.59\%		\\

			YA \cite{Haffari2012Analysis}  &57.98\% 				& 		\textbf{64.80}\%		& 69.10\%	 \\

			DSSE \cite{weston2012deep}&52.22\% 	    			          							& 		63.48\%		& 69.24\%		\\
			
			\hline
			
			DCS  									&59.12\%						     							&	64.18\%		& \textbf{69.86}\%			\\

			\hline
			
			\hline
			
		\end{tabular}
		
	\end{center}
	
\end{table}  

\begin{equation}
\label{3}
\begin{aligned}
R(\mathbf{x};\theta_t) &= \frac{\mathbf{y}_{\theta_{t-1}}(\mathbf{x})^T \mathbf{y}_{\theta_{t}}(\mathbf{x})}{|(\mathbf{y}_{\theta_{t-1}}(\mathbf{x}))^T|  |\mathbf{y}_{\theta_{t}}(\mathbf{x})|}\\ 
\end{aligned}
\end{equation}

\begin{equation}
\label{4}
\begin{aligned}
L(\mathbf{x};\theta_t) &= \underset{y}{\mathbf{argmax}}\{\mathbf{y}_{\theta_{t-1}}(\mathbf{x})\circ\mathbf{y}_{\theta_{t}}(\mathbf{x})\} \\
\end{aligned}
\end{equation}
Specifically, the feature transformation matrix $M_{f_{\theta}}(F_{t}(\mathbf{x}))$ is replaced by identity matrix, meaning the new criterion with Eq.~(\ref{3}) and Eq.~(\ref{4}) chooses unlabeled samples only considering the consistency of soft labels in transformation.

The result has been demonstrated in Fig.~\ref{ca}. There seems no distinction between the two criteria at the first iteration. Both strategies achieve high accuracy in label prediction of selected samples, and the transformation matrix merely enhances the accuracy about $1\%$. But in the second iteration, the prediction accuracy in accordance with Eq.~(\ref{3}) rapidly decreases to $68\%$; and drastically falls down to $19\%$ at the fourth iteration. In comparison, the DCS regularized by Eq.~(\ref{1}) remains accuracy above $80\%$ till the fifth iteration. The phenomenon illustrates transformation matrix $M_{f_{\theta}}(F_{t}(\mathbf{x}))$ helps to maintain the selection quality and defer the semantic drift problem.

\section{Conclusion}
\label{conclusion}
This paper presents a novel semi-supervised learning framework named Deep Co-Space (DCS) to improve deep visual classification performance via an incrementally cost-effective manner. Considering deep feature learning as a sequence of steps pursuing feature transformation, DCS proposes to measure the reliability of each unlabeled image instance by calculating the category variations of the instance and its nearest neighbors from two different neighborhood variation perspectives, and merged them into an unified sample mining criterion deriving from Hellinger distance. Extensive experiments on standard image classification benchmarks demonstrate the effectiveness of the proposed DCS. In the future, we will pay more attention to extend our DCS to other vision tasks (e.g.,object detection and segmentation).

{
\bibliographystyle{unsrt}
\bibliography{referencefinal}
}

%% biography section
%% 
%% If you have an EPS/PDF photo (graphicx package needed) extra braces are
%% needed around the contents of the optional argument to biography to prevent
%% the LaTeX parser from getting confused when it sees the complicated
%% \includegraphics command within an optional argument. (You could create
%% your own custom macro containing the \includegraphics command to make things
%% simpler here.)

%%\begin{IEEEbiography}[{\includegraphics[width=1in,height=1.25in,clip,keepaspectratio]{mshell}}]{Michael Shell}
%% or if you just want to reserve a space for a photo:
%

\begin{IEEEbiography}[{\includegraphics[width=1in,height=1.25in,clip,keepaspectratio]{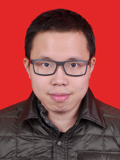}}]{Ziliang Chen} received the BS degree in  Mathematics and Applied Mathematics from Sun Yat-Sen University, Guangzhou, China. He is currently pursuing the Ph.D. degree in computer science and technology at Sun Yat-Sen University, advised by Professor Liang Lin. His current research interests include computer vision and machine learning.
\end{IEEEbiography}
\begin{IEEEbiography}[{\includegraphics[width=1in,height=1.25in,clip,keepaspectratio]{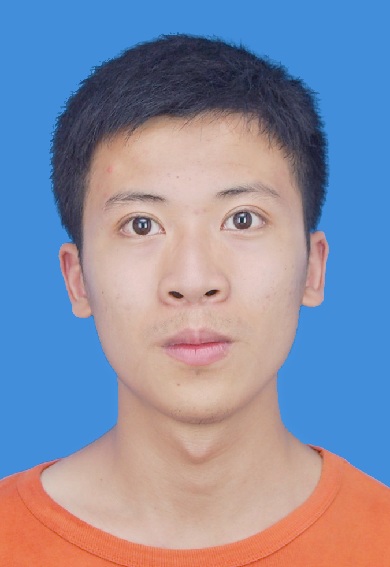}}]{Keze Wang} received his B.S. degree in software engineering from Sun Yat-Sen University, Guangzhou, China, in 2012. He is currently pursuing the dual Ph.D. degree at Sun Yat-Sen University and Hong Kong Polytechnic University, advised by Prof. Liang Lin and Lei Zhang . His current research interests include computer vision and machine learning. More information can be found in his personal website http://kezewang.com
\end{IEEEbiography}

\begin{IEEEbiography}
[{\includegraphics[width=1in,height=1.25in,clip,keepaspectratio]{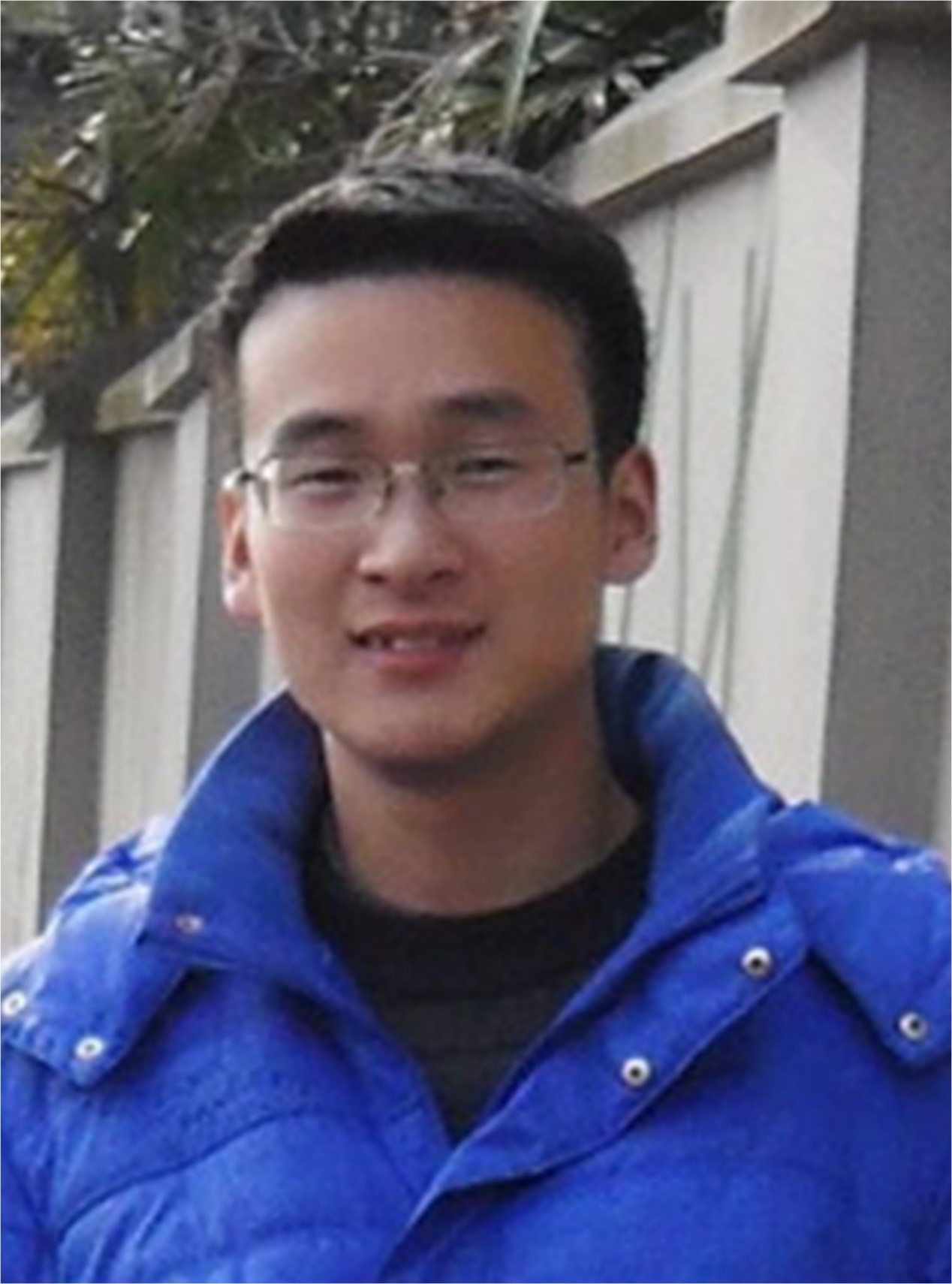}}]{Xiao Wang} received the B.S. degree in Western Anhui University, Luan, China, in 2013 and the M.S. degree in Anhui University, Hefei, China in 2016, respectively, where he is currently pursuing the Ph.D. degree in computer science in Anhui
University. His current research interests mainly about computer vision, machine learning, and pattern recognition. 
\end{IEEEbiography}

\begin{IEEEbiography}
[{\includegraphics[width=1in,height=1.25in,clip,keepaspectratio]{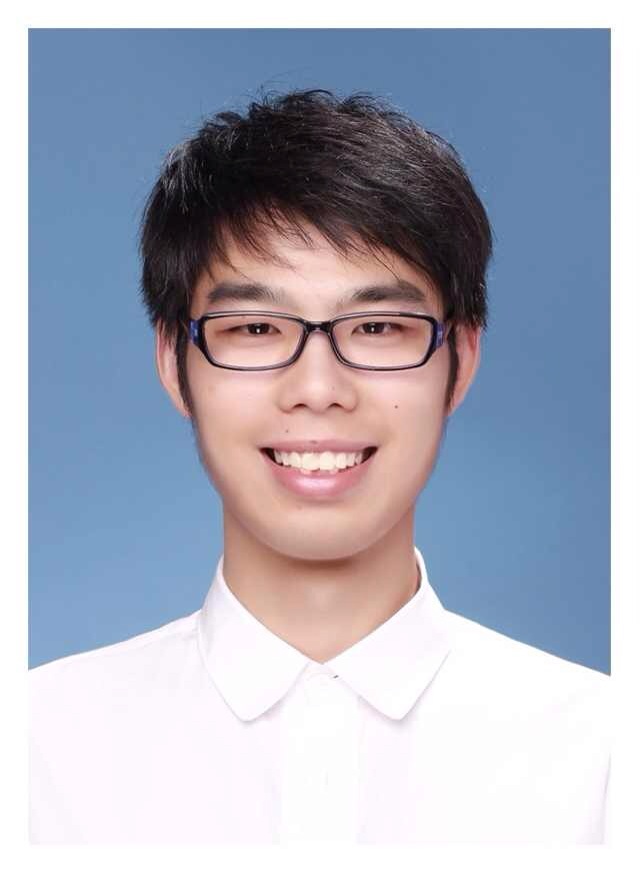}}]{Pai Peng} received the PhD degree in computer science from Zhejiang University in 2016. He is currently a research scientist in Youtu Lab of Tencent Technology (Shanghai) Co.,Ltd. His research interests include image recognition and deep learning and has published several top-tier conference and journal papers related with image recognition, e.g. SIGIR, CIKM, TKDE, ICMR, etc.
\end{IEEEbiography}
\begin{IEEEbiography}[{\includegraphics[width=1in,height=1.25in,clip,keepaspectratio]{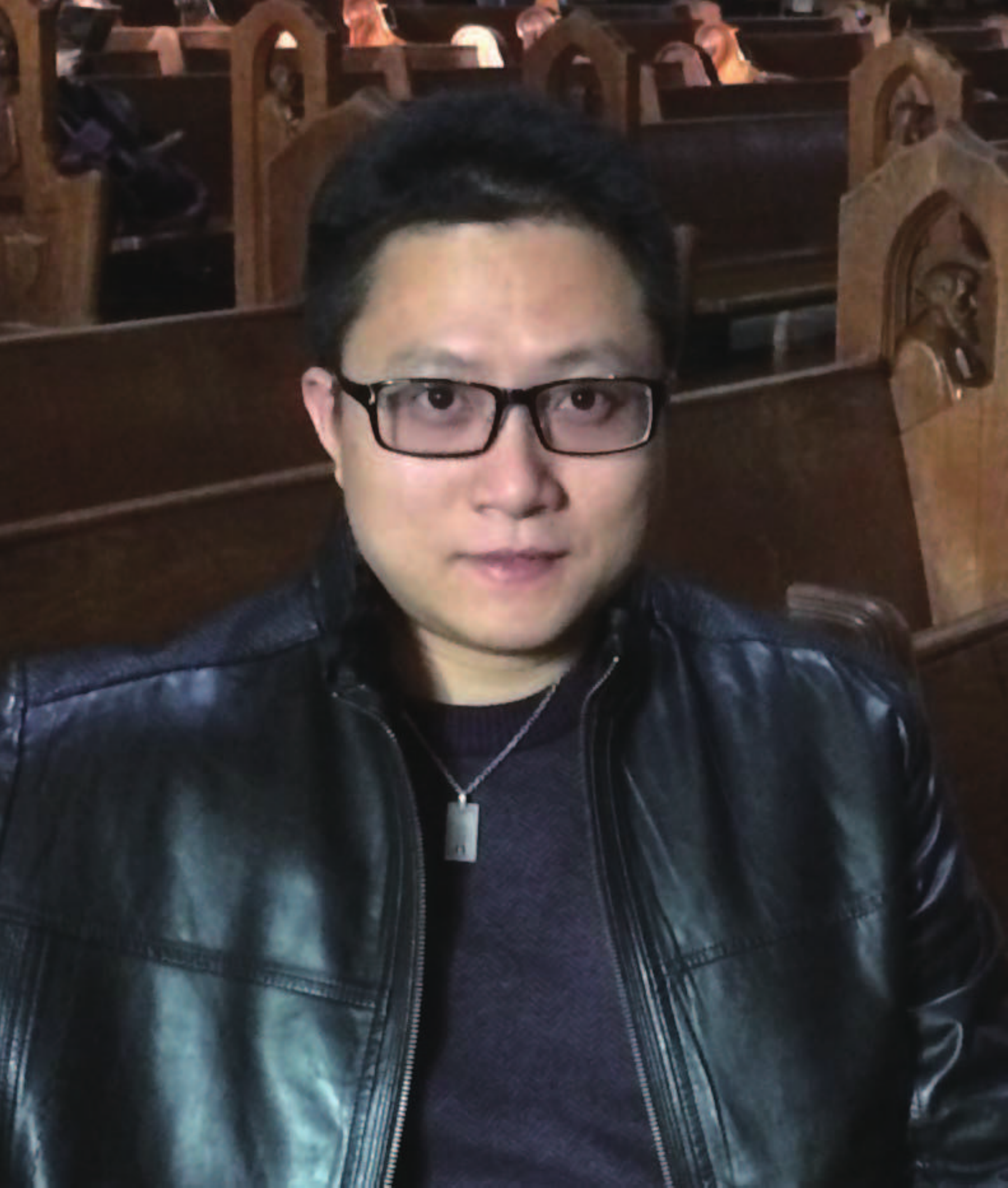}}]{Liang Lin} is a full Professor of Sun Yat-sen University. He is the Excellent Young Scientist of the National Natural Science Foundation of China. He received his B.S. and Ph.D. degrees from the Beijing Institute of Technology (BIT), Beijing, China, in 2003 and 2008, respectively, and was a joint Ph.D. student with the Department of Statistics, University of California, Los Angeles (UCLA). From 2008 to 2010, he was a Post-Doctoral Fellow at UCLA. From 2014 to 2015, as a senior visiting scholar he was with The Hong Kong Polytechnic University and The Chinese University of Hong Kong. His research interests include Computer Vision, Data Analysis and Mining, and Intelligent Robotic Systems, etc. Dr. Lin has authorized and co-authorized on more than 100 papers in top-tier academic journals and conferences. He has been serving as an associate editor of IEEE Trans. Human-Machine Systems. He was the recipient of the Best Paper Runners-Up Award in ACM NPAR 2010, Google Faculty Award in 2012, Best Student Paper Award in IEEE ICME 2014, Hong Kong Scholars Award in 2014 and Best Paper Award in IEEE ICME 2017. More information can be found in his group website http://hcp.sysu.edu.cn
\end{IEEEbiography}

%\begin{IEEEbiographynophoto}{Xiao Wang}

%% insert where needed to balance the two columns on the last page with
%%\newpage


%\vfill


\end{document}